%% file: main.tex
\documentclass[runningheads]{llncs}

\input{macro.tex}

\begin{document}
\pagestyle{headings}

\mainmatter

\title{Federated Visual Classification with Real-World Data Distribution}

\titlerunning{Real-World Federated Visual Classification}
\author{%
    Tzu-Ming Harry Hsu\inst{1,2}\thanks{work done whilst at Google.} \and
    Hang Qi\inst{2} \and
    Matthew Brown\inst{2}}
    \institute{%
    MIT CSAIL \\
    \email{stmharry@mit.edu}
    \and
    Google Research \\
    \email{\{stmharry,hangqi,mtbr\}@google.com}}

\authorrunning{T.M. Hsu et al.}

\maketitle

\input{sections/1_introduction}

\input{sections/2_related_work}

\input{sections/3_benchmark_datasets}

\input{sections/4_methods}

\input{sections/5_exp_nonid}

\input{sections/6_exp_benchmarks}
\input{sections/7_conclusions}

\bibliographystyle{splncs04}
\bibliography{main}
\clearpage

\section*{Appendix}
\input{sections/_appendix}

\end{document}

%% file: macro.tex
\usepackage[ruled,noline,noend]{algorithm2e}
\usepackage{amsfonts}
\usepackage{amsmath}
\usepackage{amssymb}
\usepackage{array}
\usepackage{bm}
\usepackage{booktabs}
\usepackage{comment}
\usepackage[T1]{fontenc}
\usepackage{graphicx}
\usepackage{hhline}
\usepackage[pagebackref=true,breaklinks=true,letterpaper=true,colorlinks,bookmarks=false]{hyperref}
\usepackage{ifthen}
\usepackage[utf8]{inputenc}
\usepackage{microtype}
\usepackage{multirow}
\usepackage{nicefrac}
\usepackage{pifont}
\usepackage{soul}
\usepackage{subcaption}
\usepackage[normalem]{ulem}
\usepackage{url}
\usepackage{xcolor}
\usepackage{xspace}
\usepackage{xstring}
\usepackage{enumerate}

\hypersetup{
    colorlinks,
    linkcolor={red},
    citecolor={green!80!black},
    urlcolor={blue}
}

\SetArgSty{textnormal}

\SetAlFnt{\footnotesize}

\SetCommentSty{xCommentSty}

\DontPrintSemicolon

\renewcommand{\subsubsection}[1]{\paragraph{\textbf{#1}}}

\newcommand{\fedavgm}{\texttt{FedAvgM}\xspace}
\newcommand{\fedavg}{\texttt{FedAvg}\xspace}
\newcommand{\fedvc}{\texttt{FedVC}\xspace}  %
\newcommand{\fedimp}{\texttt{FedIR}\xspace}  %

\newcommand{\inatgeo}[1]{Geo-#1\xspace}
\newcommand{\inatuser}{User-120k\xspace}
\newcommand{\LandmarksFed}{Landmarks-User-160k\xspace}

\newcommand{\ie}{i.e.\ }
\newcommand{\eg}{e.g.\ }
\newcommand{\etc}{etc\xspace}
\newcommand{\etal}{et al.\xspace}
\newcommand{\aka}{a.k.a.\ }
\newcommand{\vc}{\mathsf{VC}}
\def\vp{\bm{p}}
\def\vq{\bm{q}}
\def\vw{\bm{\theta}}
\def\vv{\bm{v}}
\def\cS{\mathcal{S}}
\def\cD{\mathcal{D}}

\newcommand{\ptarget}{p}

\newcommand{\vgbar}{\bar{\bm{g}}}

\def\dist{\textsf{Dist}\xspace}
\def\emd{\textsf{EMD}\xspace}

\def\dir{\mathsf{Dir}}

\def\eff{\mathrm{eff}}

\newcommand{\abs}[1]{\left|#1\right|}
\newcommand{\norm}[1]{\left\lVert#1\right\rVert}
\newcommand{\para}[1]{\left(#1\right)}
\newcommand{\param}[1]{\left[#1\right]}
\newcommand{\paral}[1]{\left\{#1\right\}}

\DeclareMathOperator{\E}{\mathbb{E}} %

\newcolumntype{L}[1]{>{\raggedright\let\newline\\\arraybackslash\hspace{0pt}}m{#1}}
\newcolumntype{C}[1]{>{\centering\let\newline\\\arraybackslash\hspace{0pt}}m{#1}}
\newcolumntype{R}[1]{>{\raggedleft\let\newline\\\arraybackslash\hspace{0pt}}m{#1}}
\newcommand{\makecell}[2][c]{\begin{tabular}[c]{@{}#1@{}}#2\end{tabular}}

\usepackage[normalem]{ulem}
\newcommand\stmharry[1]{{\color{blue}[stmharry: #1]}}
\newcommand\hq[1]{{\color{orange}[hq: #1]}}
\newcommand\matt[1]{{\color{purple}[matt: #1]}}
\newcommand\todo[1]{{\color{red}[TODO: #1]}}

\newcommand\cut[1]{}

%% file: sections/1_introduction.tex
\begin{abstract}
Federated Learning enables visual models to be trained on-device, bringing advantages for user privacy (data need never leave the device), but challenges in terms of data diversity and quality. Whilst typical models in the datacenter are trained using data that are independent and identically distributed (IID), data at source are typically far from IID. Furthermore, differing quantities of data are typically available at each device (imbalance). In this work, we characterize the effect these real-world data distributions have on distributed learning, using as a benchmark the standard Federated Averaging (FedAvg) algorithm. To do so, we introduce two new large-scale datasets for species and landmark classification, with realistic per-user data splits that simulate real-world edge learning scenarios. We also develop two new algorithms (FedVC, FedIR) that intelligently resample and reweight over the client pool, bringing large improvements in accuracy and stability in training. The datasets are made available online.

\end{abstract}

\section{Introduction}
Federated learning (FL) is a privacy-preserving framework, originally introduced by McMahan \etal~\cite{mcmahan2017communication}, for training models from decentralized user data residing on devices at the edge. Models are trained iteratively across many federated rounds. For each round, every participating device (\aka\emph{client}), receives an initial model from a central server, performs stochastic gradient descent (SGD) on its local training data and sends back the gradients. The server then aggregates all gradients from the participating clients and updates the starting model. FL preserves user privacy in that the raw data used for training models never leave the devices throughout the process. In addition, differential privacy~\cite{mcmahan2017learning} can be applied for a theoretically bounded guarantee that no information about individuals can be derived from the aggregated values on the central server.

Federated learning is an active area of research with a number of open questions~\cite{li2019federated,kairouz2019advances} remaining to be answered. A particular challenge is the distribution of data at user devices. Whilst in centralized training, data can be assumed to be independent and identically distributed (IID), this assumption is unlikely to hold in federated settings. Decentralized training data on end-user devices will vary due to user-specific habits, preferences, geographic locations, \etc. Furthermore, in contrast to the streamed batches from a central data store in the data center, devices participating in an FL round will have differing amounts of data available for training. 

In this work, we study the effect these heterogeneous client data distributions have on learning visual models in a federated setting, and propose novel techniques for more effective and efficient federated learning. We focus in particular on two types of distribution shift: \textbf{Non-Identical Class Distribution}, meaning that the distribution of visual classes at each device is different, and \textbf{Imbalanced Client Sizes}, meaning that the number of data available for training at each device varies. Our key contributions are:

\begin{itemize}
\setlength{\itemsep}{0pt}
\setlength{\parskip}{0pt}
\setlength{\parsep}{0pt}
\item \textbf{We analyze the effect of learning with per-user data} in real-world datasets, in addition to carefully controlled setups with parametric (Dirichlet) and natural (geographic) distributions.
\item \textbf{We propose two new algorithms} to mitigate per-client distribution shift and imbalance, substantially improving classification accuracy and stability.
\item \textbf{We provide new large-scale datasets} with per-user data for two classification problems (natural world and landmark recognition) to the community.
\end{itemize}

Ours is the first work to our knowledge that attempts to train large-scale visual classification models for real-world problems in a federated setting. 
We expect that more is to be done to achieve robust performance in this and related settings, and are making our datasets available to the community to enable future research in this area\footnote{\scriptsize \url{https://github.com/google-research/google-research/tree/master/federated_vision_datasets}}.

%% file: sections/2_related_work.tex
\section{Related Work}
\label{sec:related}

\subsubsection{Synthetic Client Data} Several authors have explored the \fedavg algorithm on synthetic non-identical client data partitions generated from image classification datasets. McMahan \etal~\cite{mcmahan2017communication} synthesize pathological non-identical user splits from the MNIST dataset, sorting training examples by class labels and partitioning into shards such that each client is assigned 2 shards. They demonstrate that \fedavg on non-identical clients still converges to 99\% accuracy, though taking more rounds than identically distributed clients. In a similar sort-and-partition manner, ~\cite{zhao2018federated,sattler2019robust} use extreme partitions of the CIFAR-10 dataset to form a population consisting of 10 clients in total. In contrast to these pathological data splits, Yurochkin \etal~\cite{yurochkin2019bayesian} and Hsu \etal~\cite{hsu2019measuring} synthesize more diverse non-identical datasets with Dirichlet priors.

\subsubsection{Realistic Datasets} Other authors look at more realistic data distributions at the client. For example, Caldas \etal~\cite{caldas2018leaf} use the Extended MNIST dataset~\cite{cohen2017emnist} split over the writers of the digits and the CelebA dataset~\cite{liu2015faceattributes} split by the celebrity on the picture. The Shakespeare and Stack Overflow datasets~\cite{tffdatasets} contain natural per-user splits of textual data using roles and online user ids, respectively. Luo \etal~\cite{luo2019real} propose a dataset containing 900 images from 26 street-level cameras, which they use to train object detectors. These datasets are however limited in size, and are not representative of data captured on user devices in a federated learning context. Our work aims to address these limitations (see Section \ref{sec:datasets}).

Variance reduction methods have been used in the federated learning literature to correct for the distribution shift caused by heterogeneous client data. Sahu \etal~\cite{sahu2018convergence} introduce a proximal term to client objectives for bounded variance. Karimireddy \etal~\cite{karimireddy2019scaffold} propose to use control variates for correcting client updates drift. Importance sampling is a classic technique for variance reduction in Monte Carlo methods~\cite{kahn1953methods,hesterberg1995weighted} and has been used widely in domain adaption literature for countering covariate and target shift~\cite{saerens2002adjusting,zhang2013domain,ngiam2018domain}. In this work, we adopt a similar idea of importance reweighting in a novel federated setting resulting in augmented client objectives. Different from the classic setting where samples are drawn from one proposal distribution which has the same support as the target, heterogeneous federated clients form multiple proposal distributions, each of which has partially common support with the target.

%% file: sections/3_benchmark_datasets.tex
\section{Federated Visual Classification Problems}
\label{sec:problems}

\input{floats/fig-inat}

Many problems in visual classification involve data that vary around the globe \cite{doersch2015makes,hays2008im2gps}. This means that the distribution of data visible to a given user device will vary, sometimes substantially. For example, user observations in the citizen scientist app iNaturalist will depend on the underlying species distribution in that region (see Figure~\ref{fig:inat-species-distribution}). Many other factors could potentially influence the data present on a device, including the interests of the user, their photography habits, etc. 
For this study we choose two problems with an underlying geographic variation to illustrate the general problem of non-identical user data, \emph{Natural Species Classification} and \emph{Landmark Recognition}.

\subsubsection{Natural Species Classification} We create a dataset and classification problem based on the iNaturalist 2017 Challenge~\cite{van2018inaturalist}, where images are contributed by a community of citizen scientists around the globe. Domain experts take pictures of natural species and provide annotations during field trips. Fine-grained visual classifiers could potentially be trained in a federated fashion with this community of citizen scientists without transferring images.

\subsubsection{Landmark Recognition} We study the problem of visual landmark recognition based on the 2019 Landmark Recognition Challenge~\cite{2019gldv2}, where the images are taken and uploaded by Wikipedia contributors. It resembles a scenario where smartphone users take photos of natural and architectural landmarks (e.g., famous buildings, monuments, mountains, etc.) while traveling. Landmark recognition models could potentially be trained via federated learning without uploading or storing private user photos at a centralized party.

Both datasets have data partitioning per user, enabling us to study a realistic federated learning scenario where labeled images were provided by the user and learning proceeds on-device. For experimentation in lab, we use a simulation engine for federated learning algorithms, similar to TensorFlow Federated~\cite{tff}.

\section{Datasets}
\label{sec:datasets}

In the following section, we describe in detail the datasets we develop and analyze key distributional statistics as a function of user and geo-location. We plan to make these datasets available to the community.

\input{floats/fig-inat-stat}

\subsection{iNaturalist-\inatuser and iNaturalist-Geo Splits}

iNaturalist-2017~\cite{van2018inaturalist} is a large scale fine-grained visual classification dataset comprised of images of natural species taken by citizen scientists. It has 579,184 training examples and 95,986 test examples covering over 5,000 classes. Images in this dataset are each associated with a fine-grained species label, a longitude-latitude coordinate where the picture was originally taken, and authorship information.

The iNaturalist-2017 training set has a very long-tailed distribution over classes as shown in Figure~\ref{fig:inat-stat-global}, while the test set is relatively uniform over classes.
While studying learning robustly with differing training and test distributions is a topic for research~\cite{van2017devil} in itself, in our federated learning benchmark, we create class-balanced training and test sets with uniform distributions. This allows us to focus on distribution variations and imbalance at the \emph{client level}, without correcting for overall domain shift between training and test sets.

To equalize the number of examples across classes, we first sort all class labels by their count and truncate tail classes with less than 100 training examples. This is then followed by subsampling per-class until all remaining classes each have 100 examples. This results in a balanced training set consisting of 1,203 classes and 120,300 examples. 
We use this class-balanced iNaturalist subset for the remainder of the paper.

The iNaturalist-2017 dataset includes user ids, which we use to partition the balanced training set into a population of 9,275 clients. We refer to this partitioning as \textbf{iNaturalist-\inatuser}. This contributor partitioning resembles a realistic scenario where images are collected per-user. 

In addition, for study the federated learning algorithms with client populations of varying levels of deviation from the global distribution, we also generate a \emph{wide range} of populations by partitioning the dataset at varying levels of granularity according to the geographic locations.

To utilize the geo-location tags, we leverage the S2 grid system, which defines a hierarchical partitioning of the planet surface. We perform an adaptive partitioning similar to~\cite{weyand2016planet}. Specifically, every S2 cell is recursively subdivided into four finer-level cells until no single cell contains more than $N_\textrm{max}$ examples. Cells ending up with less than $N_\textrm{min}$ examples are discarded. With this scheme, we are able to control the granularity of the resulting S2 cells such that a smaller $N_\textrm{max}$ results in a larger client count. We use $N_\textrm{max} \in$ \{30k, 3k, 1k, 100\}, $N_\textrm{min} = 0.01 N_\textrm{max}$ and refer to the resulting data partitionings as \textbf{iNaturalist-\inatgeo{\{30k, 3k, 1k, 100\}}}, respectively. Rank statistics of our geo- and per-user data splits are shown in Figures~\ref{fig:inat-stat-split-class} and~\ref{fig:inat-stat-split-example}.

\input{floats/fig-landmarks-stat}

\subsection{\LandmarksFed}

Google Landmarks Dataset V2 (GLD-v2)~\cite{2019gldv2} is a large scale image dataset for landmark recognition and retrieval, consisting of 5 million images with each attributed to one of over 280,000 authors. The full dataset is noisy: images with the same label could depict landmark exteriors, historical artifacts, paintings or sculptures inside a building. For benchmarking federated learning algorithms on a well-defined image classification problem, we use the cleaned subset (GLD-v2-clean), which is a half the size of the full dataset. In this set, images are discarded if the computed local geometric features from which cannot be matched to at least two other images with the same label~\cite{ozaki2019large}.

For creating a dataset for federated learning with natural user identities, we partition the GLD-v2-clean subset according to the authorship attribute. In addition, we mitigate the long tail while maintaining realism by requiring every landmark to have at least 30 images and be visited by at least 10 users, meanwhile requiring every user to have contributed at least 30 images that depict 5 or more landmarks. The resulting dataset has 164,172 images of 2,028 landmarks from 1,262 users, which we refer to as the train split of \textbf{\LandmarksFed}.

Following the dataset curation in \cite{2019gldv2}, the test split is created from the leftover images in GLD-v2-clean whose authors do not overlap with those in the train split. The test split contains 19,526 images and is well-balanced among classes. 1,835 of the landmarks have exactly 10 test images, and there is a short tail for the rest of the landmarks due to insufficient samples (Figure~\ref{fig:landmarks-stat}).

%% file: floats/fig-inat.tex
\begin{figure}[t]
    \centering
    \includegraphics[width=0.9\textwidth]{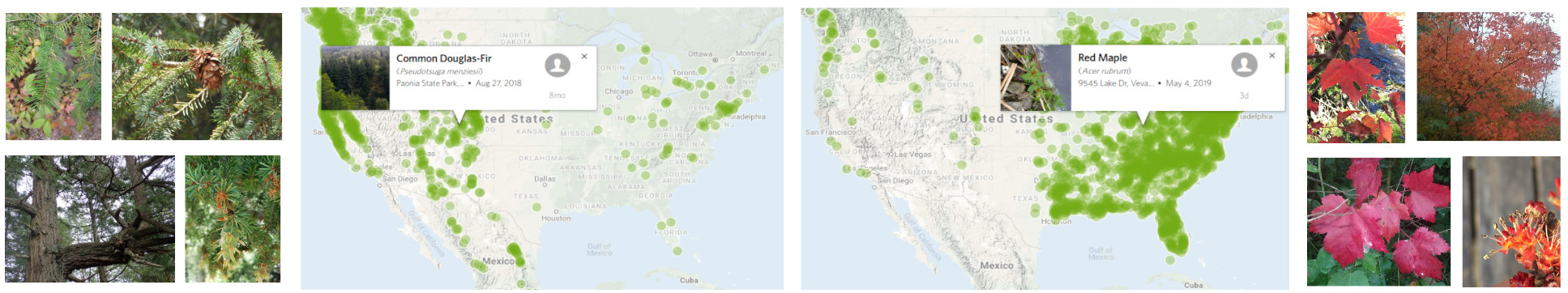}
    \caption{\textbf{iNaturalist Species Distribution}. 
    Visualized here are the distributions of Douglas-Fir and Red Maple in the continental US within iNaturalist. In a federated learning context, visual categories vary with location, and users in different locations will have very different training data distributions.
}
    \label{fig:inat-species-distribution}
\end{figure}

%% file: floats/fig-inat-stat.tex
\begin{figure}[t]
    \centering
    \begin{subfigure}[t]{0.3\linewidth}
        \centering
        \includegraphics[width=\linewidth]{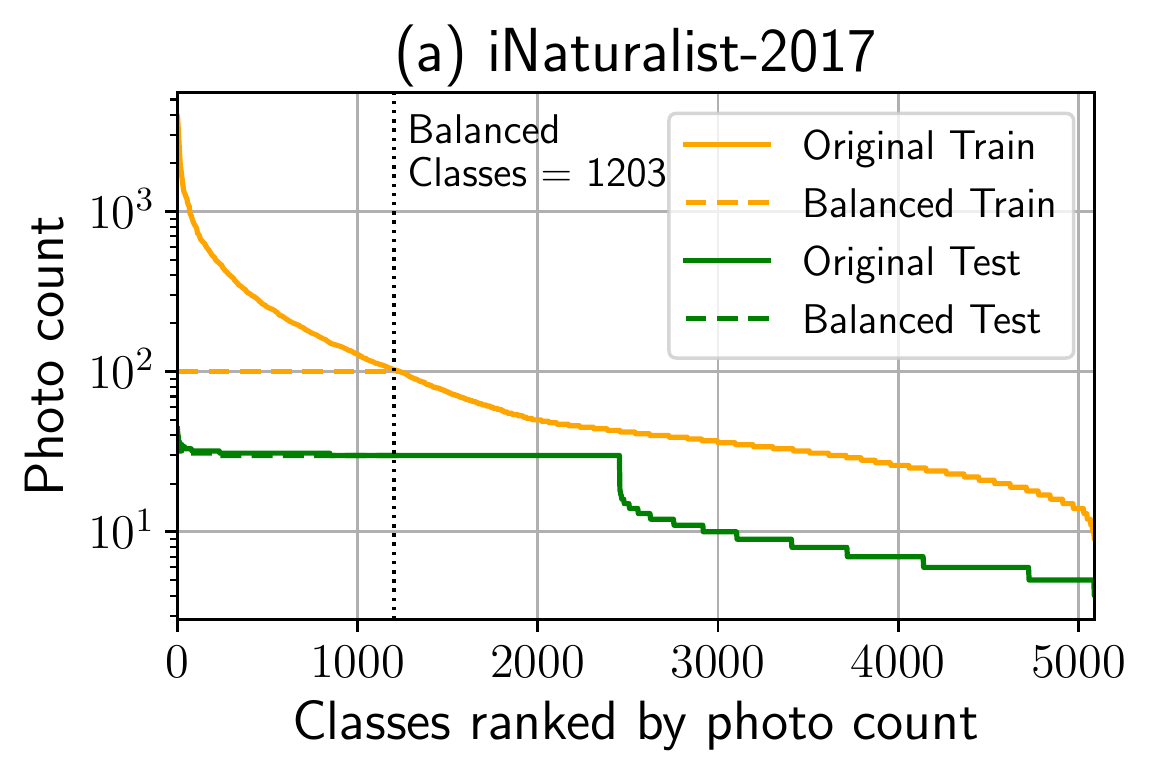}
        \phantomcaption
        \label{fig:inat-stat-global}
    \end{subfigure}%
    \begin{subfigure}[t]{0.6\linewidth}
        \centering
        \includegraphics[width=\linewidth]{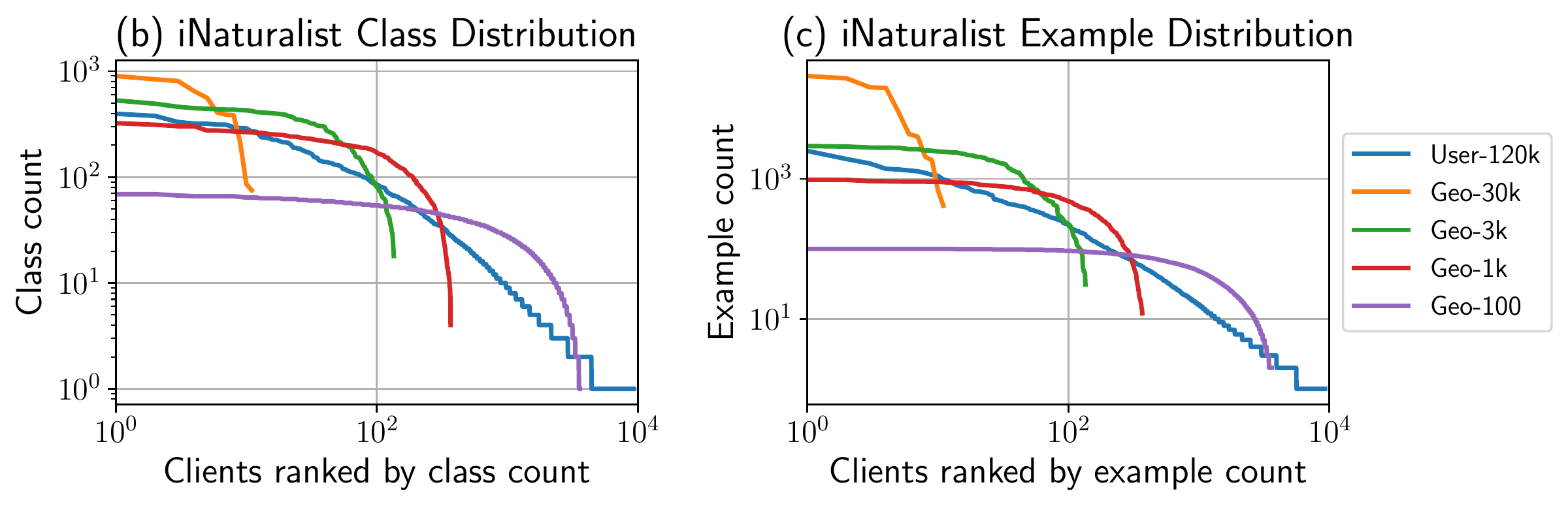}
        \phantomcaption
        \label{fig:inat-stat-split-class}
    \end{subfigure}
    \begin{subfigure}[t]{0\linewidth}
        \phantomcaption
        \label{fig:inat-stat-split-example}
    \end{subfigure}
    \caption{\textbf{iNaturalist Distribution.} In (a) we show the re-balancing of the original iNaturalist-2017 dataset. In (b) and (c) we show class and example counts vs clients for our 5 iNaturalist partitionings with varying levels of class distribution shift and size imbalance. The client count is different in each partitioning. 
    }
    \label{fig:inat-stat}
\end{figure}

%% file: floats/fig-landmarks-stat.tex
\begin{figure}[t]
    \centering
    \begin{subfigure}[t]{0pt}
        \centering
        \phantomcaption
        \label{fig:landmarks-stat-a}
    \end{subfigure}%
    \begin{subfigure}[t]{0pt}
        \centering
        \phantomcaption
        \label{fig:landmarks-stat-b}
    \end{subfigure}%
    \begin{subfigure}[t]{0pt}
        \centering
        \phantomcaption
        \label{fig:landmarks-stat-c}
    \end{subfigure}%
    \includegraphics[width=0.9\textwidth]{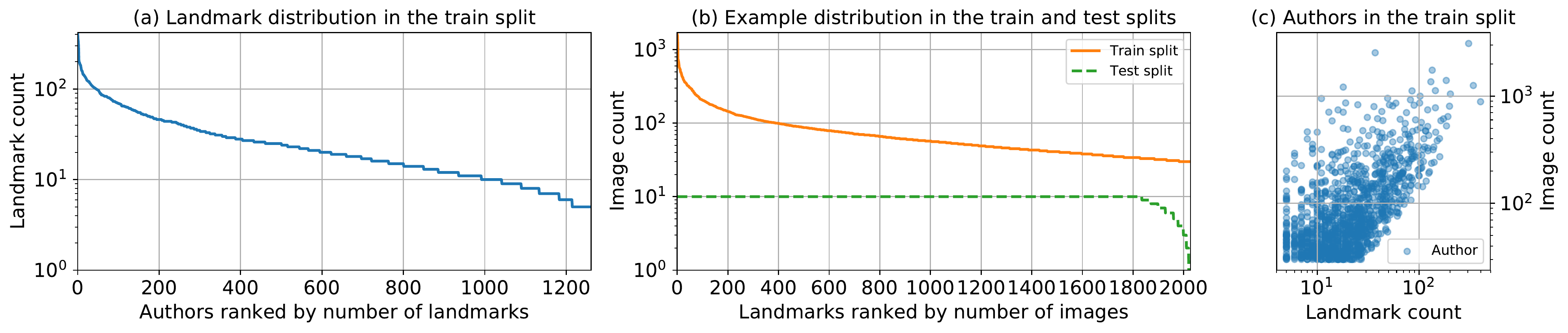}
    \includegraphics[width=0.9\textwidth]{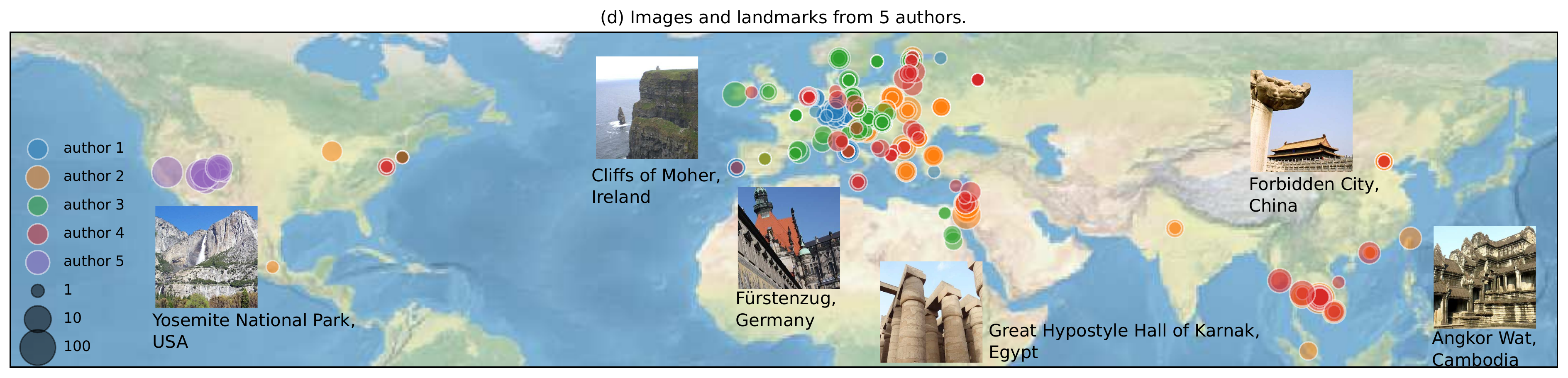}
    \caption{\textbf{\LandmarksFed Distribution}. Images are partitioned according to the authorship attribute from the GLD-v2 dataset. Filtering is applied to mitigate long tail in the train split.}
    \label{fig:landmarks-stat}
\end{figure}

%% file: sections/4_methods.tex
\section{Methods}
\label{sec:methods}

The datasets described above contain significant distribution variations among clients, which presents considerable challenges for efficient federated learning~\cite{li2019federated,kairouz2019advances}.
In the following, we describe our baseline approach of Federated Averaging algorithm (\fedavg) (Section~\ref{sec:fedavg}) and two new algorithms intended to specifically address the non-identical class distributions and imbalanced client sizes present in the data (Sections~\ref{sec:importance-sampling} and~\ref{sec:fedvc} respectively).

\input{floats/alg-fedavg}

\subsection{Federated Averaging and Server Momentum}
\label{sec:fedavg}

A standard algorithm~\cite{mcmahan2017communication} for FL, and the baseline approach used in this work, is Federated Averaging (\fedavg). See Algorithm~\ref{alg:fedavg-AllInOne}. For every federated round, $K$ clients (the \emph{report goal}) are randomly selected with uniform probability from a pool of active clients. Selected clients, indexed by $k$, download the same starting model $\vw_t$ from a central server and perform local SGD optimization, minimizing an empirical loss $L(b)$ over local mini-batches $b$ with learning rate $\eta$, for $E$ epochs before sending the accumulated model update $\Delta \vw_t^k$ back to the server. The server then averages the updates from the reporting clients $\vgbar_t = \sum_{k=1}^{K} \frac{n_k}{n} \Delta \vw_t^k$ with weights proportional to the sizes of clients' local data and finishes the federated round by applying aggregated updates to the starting model $\vw_{t+1} \leftarrow \vw_t - \gamma \vgbar_t$, where $\gamma$ is the server learning rate. Given this framework, alternative optimizers can be applied. \fedavgm~\cite{hsu2019measuring} has been shown to improve robustness to non-identically distributed client data. It uses a momentum optimizer on the server with the update rule $\vw_{t+1} \leftarrow \vw_t - \gamma \vv_t$, where $\vv_t \leftarrow \beta \vv_{t-1} + \vgbar_t$ is the exponentially weighted moving average of the model updates with powers of $\beta$.

\subsection{Importance Reweighted Client Objectives}
\label{sec:importance-sampling}

Now we address the non-identical class distribution shift in federated clients. Importance reweighting are commonly used for learning from datasets distributed differently from a target distribution. Given that the distribution variation among clients is an inherent characteristic of FL, we propose the following scheme.

Consider a target distribution $\ptarget(x, y)$ of images $x$ and class labels $y$ on which a model is supposed to perform well (\eg a validation dataset known to the central server), and a predefined loss function $\ell(x, y)$. The objective of learning is to minimize the expected loss $\E_{\ptarget}[\ell(x, y)]$ with respect to the target distribution $\ptarget$. SGD in the centralized setting achieves this by minimizing an empirical loss on mini-batches of IID training examples from the same distribution, which are absent in the federated setting. Instead, training examples on a federated client $k$ are sampled from a client-specific distribution $q_k(x, y)$. This implies that the empirical loss being optimized on every client is a \textit{biased} estimator of the loss with respect to the target distribution, since $\E_{q_k}[\ell(x, y)] \neq \E_{\ptarget}[\ell(x, y)]$.

We propose an importance reweighting scheme, denoted \fedimp, that applies importance weights $w_k(x, y)$ to every client's local objective as follows
\begin{equation}
\small
\tilde{\ell}(x, y) = \ell(x, y) w_k(x, y),
\textrm{\quad where }
w_k(x, y) = \frac{\ptarget(x,y)}{q_k(x,y)}.
\end{equation}
With the importance weights in place, an unbiased estimator of loss with respect to the target distribution can be obtained using training examples from the client distribution
\begin{equation}
\small
\E_{\ptarget}\left[\ell(x, y)\right] = \sum_{x,y} \dfrac{ \ell(x,y)\ptarget(x,y)}{q_k(x,y)}q_k(x,y) = \E_{q_k}\param{\ell(x,y) \frac{\ptarget(x,y)}{q_k(x,y)}}.
\end{equation}

Assuming that all clients share the same conditional distribution of images given a class label as the target, i.e. $\ptarget(x|y) \approx q_k(x|y)\ \forall k$, the importance weights can be computed on every client directly from the class probability ratio
\begin{equation}
\small
w_k(x, y) = \frac{\ptarget(x,y)}{q_k(x,y)} = \frac{\ptarget(y)\ptarget(x|y)}{q_k(y)q_k(x|y)} \approx \frac{\ptarget(y)}{q_k(y)}.
\end{equation} %
Note that this computation does not sabotage the privacy-preserving property of federated learning. The denominator $q_k(y)$ is private information available locally at and never leaves client $k$, whereas the numerator $\ptarget(y)$ does not contain private information about clients and can be transmitted from the central server with minimal communication cost: $C$ scalars in total for $C$ classes.

Since scaling the loss also changes the effective learning rate in the SGD optimization, in practice, we use self-normalized weights when computing loss over a mini-batch $b$
\begin{equation}
\small
\label{eq:fedimp-loss}
\tilde{L}(b) = \dfrac{\sum_{\para{x, y} \in b} \ell(x, y) w_k(x, y)}{\sum_{\para{x, y} \in b} w_k(x, y)}.
\end{equation} %
This corresponds to the self-normalized importance sampling in the statistics literature~\cite{hesterberg1995weighted}. \fedimp does not change server optimization loops and can be applied together with other methods, such as \fedavgm. See Algorithm~\ref{alg:fedavg-AllInOne}.

\subsection{Splitting Imbalanced Clients with Virtual Clients}
\label{sec:fedvc}

The number of training examples in users' devices vary in the real world.
Imbalanced clients can cause challenges for both optimization and engineering practice. Previous empirical studies~\cite{mcmahan2017communication,hsu2019measuring} suggest that the number of local epochs $E$ at every client has crucial effects on the convergence of \fedavg. A larger $E$ implies more optimization steps towards local objectives being taken, which leads to slow convergence or divergence due to increased variance. Imbalanced clients suffer from this optimization challenge even when $E$ is small. Specifically, a client with a large number of training examples takes significantly more local optimization steps than another with fewer training examples. This difference in steps is proportional to the difference in the number of training examples. In addition, a client with an overly large training dataset will take a long time to compute updates, creating a bottleneck in the federated learning round. Such clients would be abandoned by a FL production system in practice, if failing to report back to the central server within a certain time window~\cite{bonawitz2019towards}.

We hence propose a new \textit{Virtual Client} (\fedvc) scheme to overcome both issues. The idea is to conceptually split large clients into multiple smaller ones, and repeat small clients multiple times such that all \textit{virtual} clients are of similar sizes. 
To realize this, we fix the number of training examples used for a federated learning round to be $N_\vc$ for every client, resulting in exactly $S = N_\vc / B$ optimization steps taken at every client given a mini-batch size $B$. 
Concretely, consider a client $k$ with a local dataset $\cD_k$ with size $n_k = \abs{\cD_k}$. A random subset consisting of $N_\vc$ examples is uniformly resampled from $\cD_k$ for every round the client is selected. This resampling is conducted without replacement when $n_k \geq N_\vc$; with replacement otherwise. In addition, to avoid underutilizing training examples from large clients, the probability that any client is selected for a round is set to be proportional to the client size $n_k$, in contrast to uniform as in \fedavg. Key changes are outlined in Algorithm~\ref{alg:fedavg-AllInOne}. It is clear that \fedvc is equivalent to \fedavg when all clients are of the same size.

%% file: floats/alg-fedavg.tex
\SetAlgoSkip{}
\SetAlgoInsideSkip{}
\begin{algorithm}[t]
\SetKwProg{Fn}{}{:}{}
\SetKwFunction{FedAvg}{FedAvg}
\SetKwFunction{SelectClients}{SelectClients}
\SetKwFunction{ClientUpdate}{ClientUpdate}
\SetKwFunction{AggregateClient}{AggregateClient}

\SetKwComment{Change}{\color{blue}$\triangleright$\ }{}
\SetKwComment{ChangeIR}{\color{red}$\triangleright$\ }{}

\caption{\fedavg, {\color{red}\fedimp}, and {\color{blue}\fedvc}.}
\label{alg:fedavg-AllInOne}
\small%

\textbf{Server training loop:} \; \Indp
Initialize $\vw_0$ \;
\For{each round $t = 0, 1, \ldots$}{
    Subset of $K$ clients $\leftarrow$ \SelectClients{$K$} \;
    \ForPar{each client $k = 1, 2, \ldots, K$}{
        $\Delta \vw_t^k \leftarrow$ \ClientUpdate{$k$, $\vw_t$} \;
    }
    $\vgbar_t \leftarrow $ \AggregateClient{$\{ \Delta \vw_t^k \}_{k=1}^K$ } \;
    $\vw_{t+1} \leftarrow \vw_t - \gamma \vgbar_t $ \;
}
\Indm
\BlankLine
\Fn{\SelectClients{$K$}}{
    \Return $K$ clients sampled uniformly  \Change*[r]{\scriptsize\color{blue}with probability $\propto n_i$ for client $i$}
}
\BlankLine
\Fn{\ClientUpdate{$k$, $\vw_t$}}{
    $\vw \leftarrow \vw_t$ \;
    \For(\Change*[f]{\scriptsize\color{blue}over $S$ steps}){each local mini-batch $b$ over $E$ epochs}{
      $\vw \leftarrow \vw - \eta\nabla L(b; \vw)$ \ChangeIR*[r]{\scriptsize\color{red}$\nabla \tilde{L}(b; \vw)$ in Eq.\ref{eq:fedimp-loss}}
    }
    \Return $\Delta \vw \leftarrow \vw_t - \vw$ to server
}
\BlankLine
\Fn{\AggregateClient{$\{ \Delta \vw_t^k \}_{k=1}^K$}}{
    \Return $\sum_{k=1}^{K} \frac{n_k}{n} \Delta \vw^k_t$, where $n = \sum_{k=1}^K n_k$ \Change*[r]{\scriptsize\color{blue}$\frac{1}{K} \sum_{k=1}^{K} \Delta \vw^k_t$}
}
\end{algorithm}

%% file: sections/5_exp_nonid.tex
\section{Experiments}

\input{floats/tbl-dataset-stats}

We now present an empirical study using the datasets and methods of Sections~\ref{sec:datasets}~and~\ref{sec:methods}. We start by analyzing the classification performance as a function of non-identical data distribution (Section~\ref{sec:acc_vs_noniid}), using the CIFAR10/100 datasets. Next we show how \emph{Importance Reweighting} can improve performance in the more non-identical cases (Section \ref{sec:importance_reweighting}). With real user data, clients are also imbalanced, we show how this can be mitigated with \emph{Federated Virtual Clients} in Section~\ref{sec:fedvc}. Finally we present a set of benchmark results with the per-user splits of iNaturalist and Landmark datasets (Section~\ref{sec:benchmarks}). A summary of the datasets used is provided in Table~\ref{tbl:dataset-stats}. Implementation details are deferred to Section~\ref{sec:implementation}.

\subsubsection{Metrics}
When using the same dataset, the performance of a model trained with federated learning algorithms is inherently upper bounded by that of a model trained in the centralized fashion. We evaluate the \emph{relative accuracy}, defined as $\textrm{Acc}_\textrm{federated}/\textrm{Acc}_\textrm{centralized}$, and compare this metric under different types of budgets. The centralized training baseline uses the same configurations and hyperparameters for a fair comparison.

\input{floats/fig-non-identicalness}

\subsection{Classification Accuracy vs Distribution Non-Identicalness}
\label{sec:acc_vs_noniid}

Our experiments use CIFAR10/100 datasets to characterize classification accuracy with a continuous range of distribution non-identicalness. We follow the protocol described in \cite{hsu2019measuring} such that the class distribution of every client is sampled from a Dirichlet distribution with varying concentration parameter $\alpha$. 

We measure distribution non-identicalness using an average \emph{Earthmover's Distance} (\emd) metric. Specifically, we take the discrete class distribution $\vq_i$ for every client, and define the population's class distribution as $\vp = \sum_i \frac{n_i}{n} \vq_i$, where $n=\sum_i n_i$ counts training samples from all clients. 
The non-identicalness of a dataset is then computed as the weighted average of distances between clients and the population: $\sum_i \frac{n_i}{n} {\dist\para{\vq_i, \vp}}$.
$\dist\para{\cdot, \cdot}$ is a distance metric between two distributions, which we, in particular, use 
$\emd\para{\vq, \vp} \equiv \norm{\vq - \vp}_1 $, bounded between $[0, 2]$.

Figures~\ref{fig:non-identicalness-cifar-10}~and~\ref{fig:non-identicalness-cifar-100} show the trend in classification accuracy as a function of distribution non-identicalness (average \emd difference). 
We are able to approach centralized learning accuracy with data on the identical end.
A substantial drop around an \emd of 1.7 to 2.0 is observed in both datasets.
Applying momentum on the server, \fedavgm significantly improves the convergence under heterogeneity conditions for all datasets. 
Using more clients per round (larger report goal $K$) is also beneficial for training but has diminishing returns.

\input{floats/fig-importance-sampling}

\subsection{Importance Reweighting}
\label{sec:importance_reweighting}
Importance Reweighting is proposed for addressing the per-client distribution shift. We evaluate \fedimp with both \fedavg and \fedavgm on both two datasets with natural user splits: iNaturalist-\inatuser and \LandmarksFed.

For Landmarks, we experiment with two different training schemes: (a) fine-tuning the entire network (\emph{all layers}) end to end, (b) only training the last \emph{two layers} while freezing the network backbone. We set the local epochs to $E=5$ and experiment with report goals $K=$ \{10, 50, 100\}, respectively.

The result in Figure~\ref{fig:importance-sampling} shows a consistent improvement on the \LandmarksFed dataset over the \fedavg baseline. 
While \fedavgm gives the most significant improvements in all runs, \fedimp further improves the convergence speed and accuracy especially when the report goal is small (Figure~\ref{fig:landmarks}). 

\LandmarksFed ($\emd=1.94$) has more skewed data distribution than iNaturalist-\inatuser ($\emd=1.83$) and benefits more from \fedimp.

\input{floats/tbl-virtual-client}
\input{floats/fig-virtual-client}

\subsection{Federated Virtual Clients}
\label{sec:fedvc_expts}

We apply the Virtual Clients scheme (\fedvc) to both \fedavg and \fedavgm and evaluate its efficacy using iNaturalist user and geo-location datasets, each of which contains significantly imbalanced clients. In the experiments, 10 clients are selected for every federated round. We use a mini-batch size $B = 64$ and set the virtual client size $N_\vc = 256$.

Figure~\ref{fig:virtual-client} demonstrates the efficiency and accuracy improvements gained via \fedvc when clients are imbalanced. The convergence of vanilla \fedavg suffers when clients perform excessive local optimization steps. In iNaturalist-\inatgeo{3k}, for example, clients can take up to 46 (\ie 3000/64) local steps before reporting to the server. To show that \fedvc utilizes data efficiently, we report accuracy at fixed batch budgets in addition to fixed round budgets. Batch budget is calculated by summing the number of local batches taken for the largest client per round. 
As shown in Table~\ref{tbl:virtual-client}, \fedvc consistently yields superior accuracy on both \fedavg and \fedavgm. Learning curves in Figure~\ref{fig:virtual-client} show that \fedvc also decreases the learning volatility and stabilizes learning.

iNaturalist per-user and geo-location datasets reflect varying degrees of non-identicalness. Figure~\ref{fig:non-identicalness-inat}, though noisier, exhibits a similar trend compared to Figure~\ref{fig:non-identicalness}. The performance degrades as the degree of non-identicalness, characterized by \emd, increases.

%% file: floats/tbl-dataset-stats.tex
\begin{table}[t]
\centering
\caption{ \textbf{Training Dataset Statistics.} Note that while CIFAR-10/100 and iNaturalist datasets each have different partitionings with different levels of identicalness, the underlying data pool is unchanged and thus sharing the same centralized learning baselines. }
\label{tbl:dataset-stats}
\scriptsize
\begin{tabular}{ L{3cm}R{1.5cm}R{1.5cm}R{1.5cm}R{1.5cm}R{1.75cm} }
\toprule
\multirow{2}{*}[-0.7em]{\textbf{ Dataset}} & \multicolumn{2}{c}{\textbf{ Clients}} & \multicolumn{1}{c}{\textbf{ Classes}} & \multicolumn{1}{c}{\textbf{ Examples}} & \multirow{2}{*}[-0.7em]{\textbf{\makecell[r]{ Centralized \\ Accuracy}}} \\ \cmidrule{2-3} \cmidrule(l){4-4} \cmidrule(l){5-5}
  & \makecell[r]{ Count} & \makecell[r]{ Size \\ Imbalance} & \makecell[r]{ Count} & \makecell[r]{ Count} &   \\ \midrule
Synthetic &   &   &   &   \\
\quad CIFAR-10 & 100 & \ding{55} & 10 & 50,000 & 86.16\% \\
\quad CIFAR-100 & 100 & \ding{55} & 100 & 50,000 & 55.21\% \\
\quad iNaturalist Geo Splits & 11 to 3606 & \ding{51} & 1,203 & 120,300 & 57.90\% \\ \midrule
Real-World &   &   &   &   \\
\quad iNaturalist-User-120k & 9,275 & \ding{51} & 1,203 & 120,300 & 57.90\% \\
\quad \LandmarksFed & 1,262 & \ding{51} & 2,028 & 164,172 & 67.05\% \\ \bottomrule
\end{tabular}
\end{table}

%% file: floats/fig-non-identicalness.tex
\begin{figure}[t]
    \centering
    \begin{subfigure}[t]{0.45\linewidth}
        \centering
        \includegraphics[width=\linewidth]{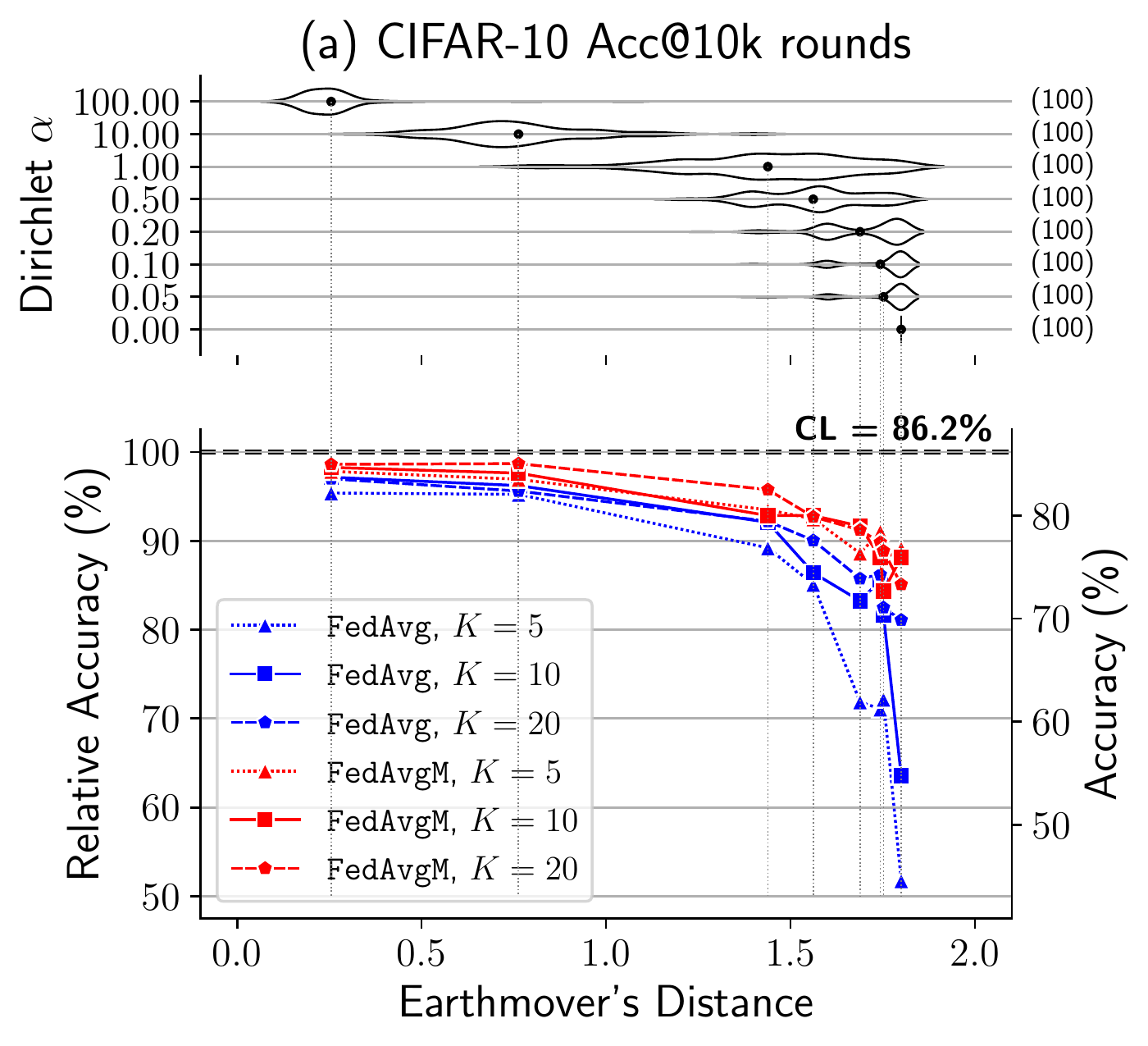}
        \phantomcaption
        \label{fig:non-identicalness-cifar-10}
    \end{subfigure}%
    ~
    \begin{subfigure}[t]{0.45\linewidth}
        \centering
        \includegraphics[width=\linewidth]{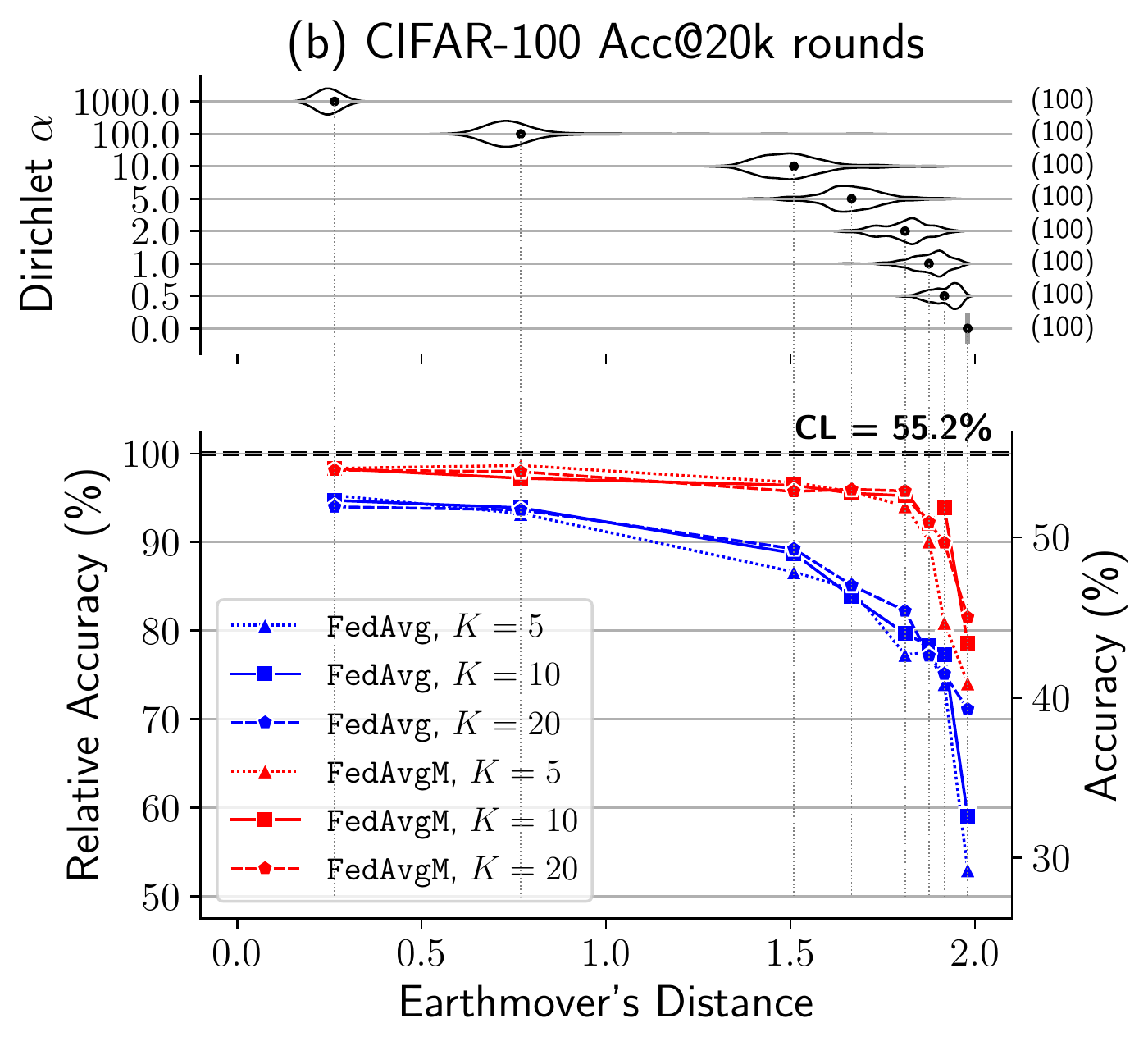}
        \phantomcaption
        \label{fig:non-identicalness-cifar-100}
    \end{subfigure}
    \caption{\textbf{Relative Accuracy v.s. Non-identicalness.} Federated learning experiments are performed on (a) CIFAR-10 and (b) CIFAR-100 using local epoch $E=1$. The top row demonstrates the distributions of \emd of clients with different data partitionings. Total client counts are annotated to the right, and the weighted average of all client \emd is marked. Data is increasingly non-identical to the right. The dashed line indicates the centralized learning performance. The best accuracies over a grid of hyperparameters are reported (see Appendix \ref{sec:hyper}). }
    \label{fig:non-identicalness}
\end{figure}

%% file: floats/fig-importance-sampling.tex
\begin{figure}[t]
    \centering
    \includegraphics[width=0.9\textwidth]{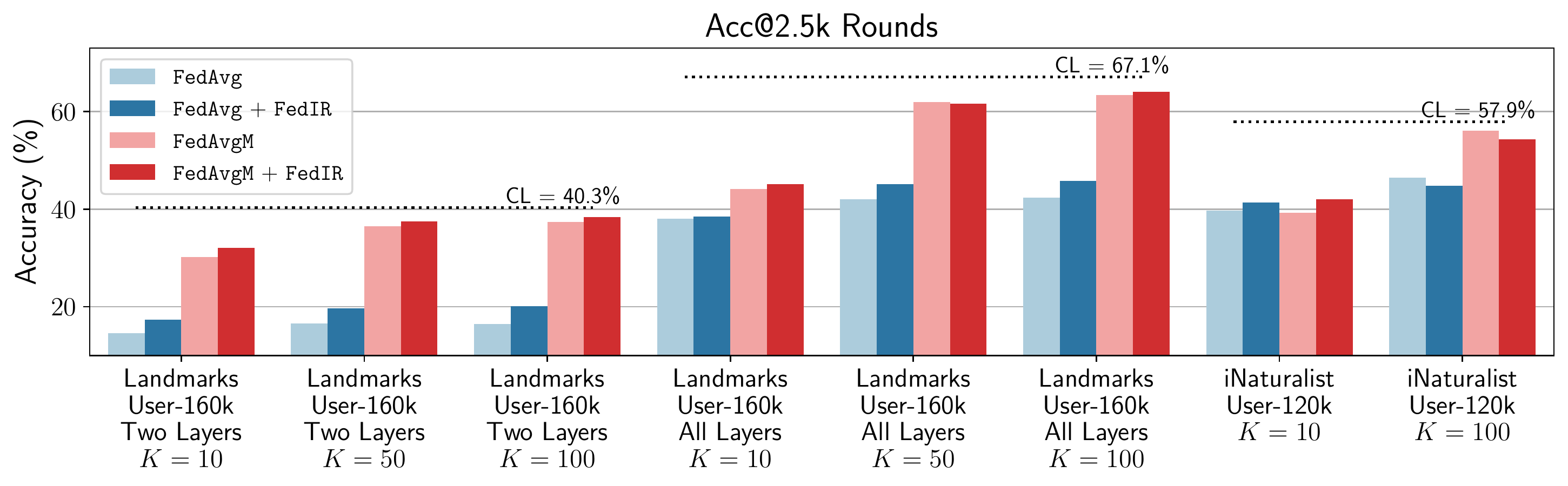}
    \caption{\textbf{Comparing Base Methods with and without \fedimp.} Accuracy shown at 2.5k communication rounds. Centralized learning accuracy marked with dashed lines.
    }
    \label{fig:importance-sampling}
\end{figure}

%% file: floats/tbl-virtual-client.tex
\begin{table}[t]
\centering
\caption{ \textbf{Accuracy of Federated Virtual Client on iNaturalist.} Acc@round denotes the accuracy at a FL communication round. Acc@batch denotes the batch count accumulated over the largest clients per round, and is a proxy for a fixed time budget. }
\label{tbl:virtual-client}
\scriptsize
\begin{tabular}{ C{1.75cm}L{1.25cm}C{1.0cm}C{1.0cm}C{1.0cm}C{1.0cm}C{1.0cm}C{1.0cm}C{1.0cm}C{1.0cm} }
\toprule
\multirow{2}{*}[-0.2em]{\textbf{ Data}} & \multirow{2}{*}[-0.2em]{\textbf{ Method}} & \multirow{2}{*}[-0.2em]{\textbf{ \fedvc}} & \multirow{2}{*}[-0.2em]{\textbf{ $K$}} & \multicolumn{3}{c}{\textbf{ Acc@Round(\%)}} & \multicolumn{3}{c}{\textbf{ Acc@Batch(\%)}} \\ \cmidrule{5-7} \cmidrule(l){8-10}
  &   &   &   & 1k & 2.5k & 5k & 10k & 25k & 50k \\ \midrule
\multirow{4}{*}[-0.3em]{\textbf{ \inatgeo{3k}}} & \fedavg & \ding{55} & 10 & 47.0 & 47.9 & 48.7 & 37.8 & 44.4 & 46.5 \\
  & \fedavgm & \ding{55} & 10 & 47.2 & 50.4 & 45.0 & 42.5 & 47.1 & 44.9 \\ \cmidrule{2-10}
  & \fedavg & \ding{51} & 10 & 37.4 & 46.2 & 52.8 & 46.2 & 53.1 & 55.5 \\
  & \fedavgm & \ding{51} & 10 & \textbf{49.7} & \textbf{54.8} & \textbf{56.7} & \textbf{54.8} & \textbf{56.7} & \textbf{57.1} \\ \midrule
\multirow{4}{*}[-0.3em]{\textbf{ \inatuser}} & \fedavg & \ding{55} & 10 & 34.7 & 39.7 & 41.3 & 37.8 & 39.8 & 42.9 \\
  & \fedavgm & \ding{55} & 10 & 31.9 & 39.2 & 41.3 & 32.3 & 41.6 & 43.4 \\ \cmidrule{2-10}
  & \fedavg & \ding{51} & 10 & 31.3 & 39.7 & 43.9 & 39.7 & \textbf{48.9} & 52.8 \\
  & \fedavgm & \ding{51} & 10 & \textbf{37.9} & \textbf{43.7} & \textbf{49.1} & \textbf{43.7} & 47.4 & \textbf{54.6} \\ \midrule
  & \multicolumn{3}{l}{ Centralized} & \multicolumn{6}{c}{ 57.9} \\ \bottomrule
\end{tabular}
\end{table}

%% file: floats/fig-virtual-client.tex
\begin{figure}[t]
    \centering
    \begin{minipage}[h]{0.4\linewidth}
        \begin{subfigure}[h]{1.0\linewidth}
            \includegraphics[width=1.0\linewidth]{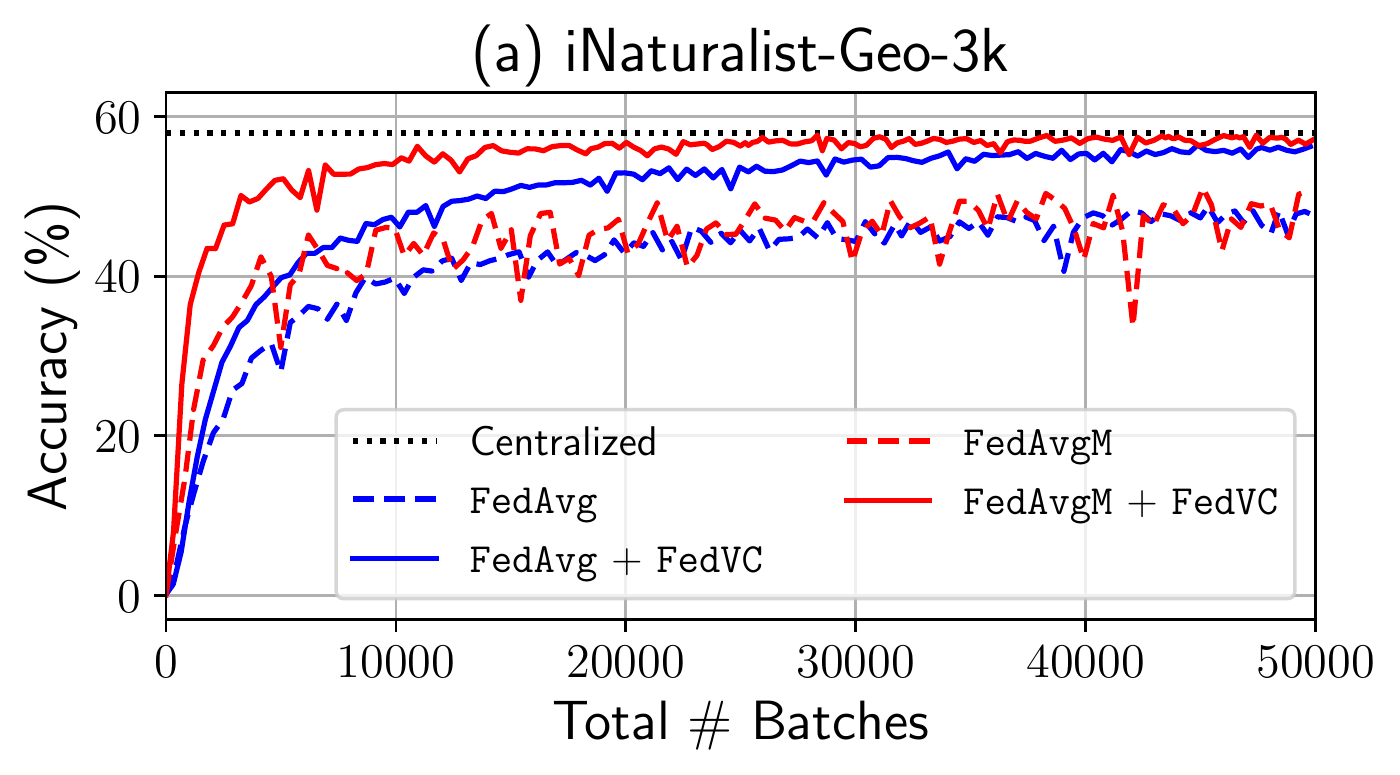}
            \phantomcaption
        \end{subfigure}
        
        \begin{subfigure}[h]{1.0\linewidth}
            \includegraphics[width=1.0\linewidth]{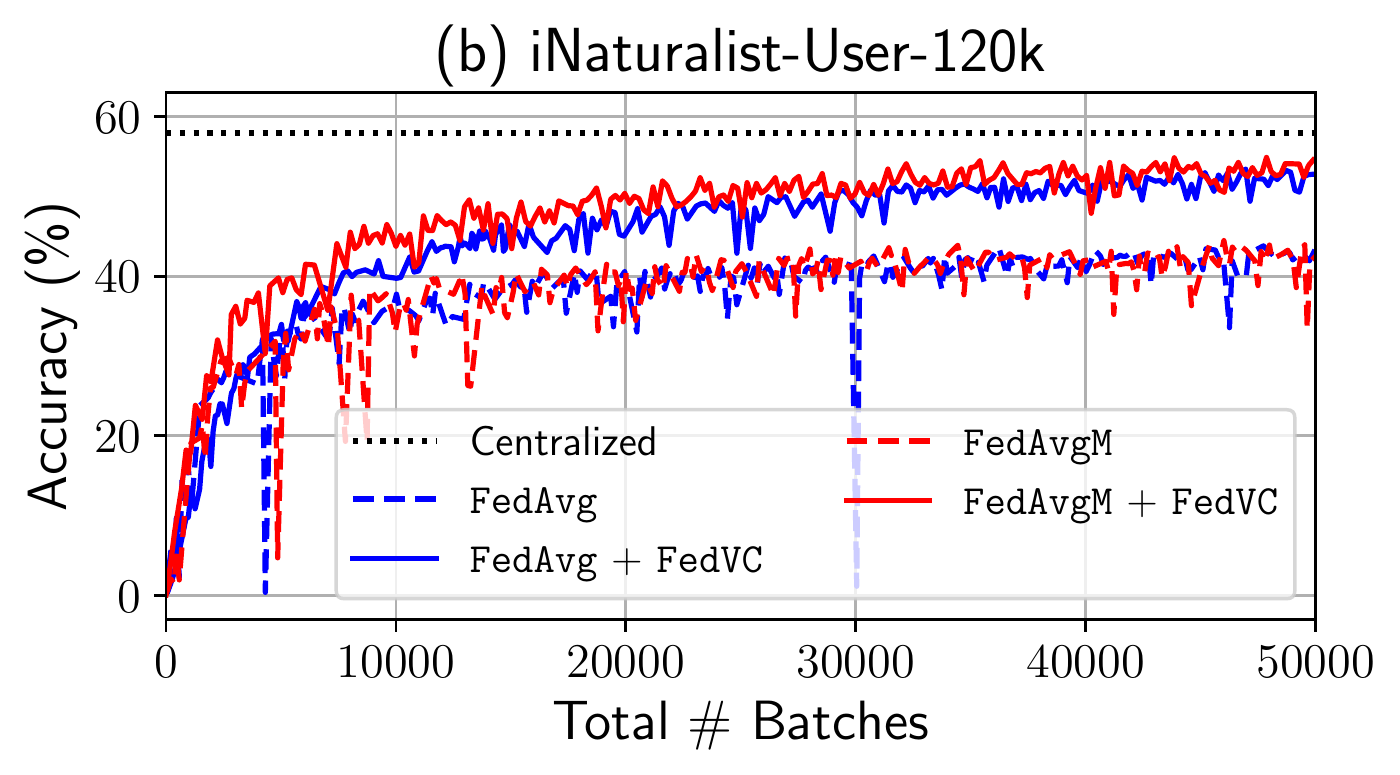}
            \phantomcaption
        \end{subfigure}
    \end{minipage}%
    ~
    \begin{minipage}[h]{0.45\linewidth}
        \begin{subfigure}[h]{1.0\linewidth}
            \centering
            \includegraphics[width=1.0\linewidth]{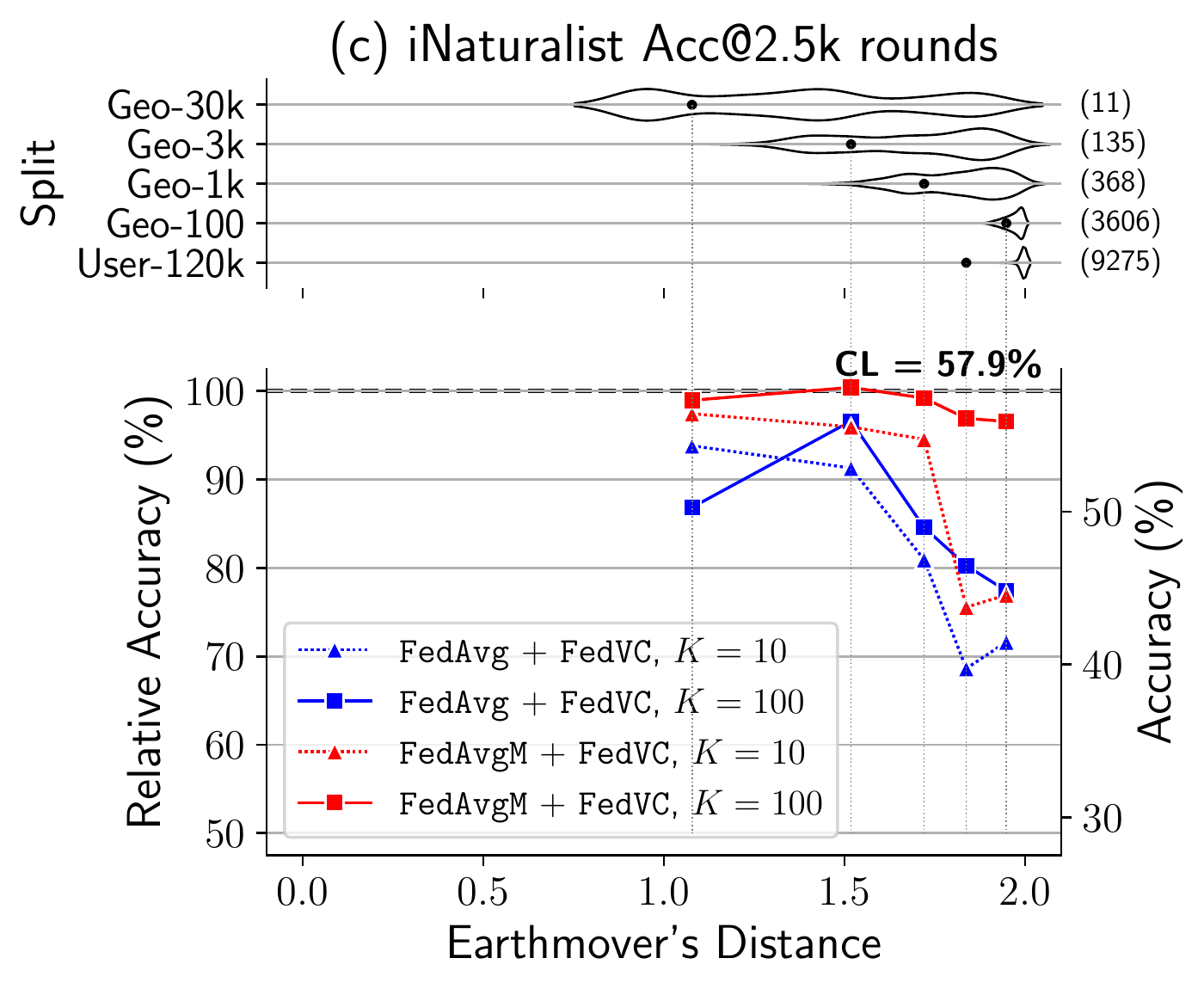}
            \phantomcaption
            \label{fig:non-identicalness-inat}
        \end{subfigure}
    \end{minipage}
    \caption{
        \textbf{Learning with Federated Virtual Clients.}
        Curves on the left are learned on the iNaturalist geo-partitioning \inatgeo{3k} and user split \inatuser each with 135 clients and 9275 clients. Experiments on multiple iNaturalist partitionings are shown on the right, plotting relative accuracy at 2.5k communication rounds to mean \emd. Centralized learning achieves a 57.9\% accuracy.}
    \label{fig:virtual-client}
\end{figure}

%% file: sections/6_exp_benchmarks.tex
\subsection{Federated Visual Classification Benchmarks}
\label{sec:benchmarks}

Having shown that our proposed modifications to \fedavg indeed lead to a speedup in learning on both iNaturalist and Landmarks, we wish to also provide some benchmark results on natural user partitioning with reasonable operating points. We hope that these datasets can be used for understanding real-world federated visual classification, and act as benchmarks for future improvements.

\input{floats/tbl-inaturalist}

\subsubsection{iNaturalist-\inatuser}
The iNaturalist-\inatuser data has 9,275 clients and 120k examples, containing 1,203 species classes. We use report goals $K=$ \{10, 100\}. \fedvc samples $N_\vc = 256$ examples per client. A summary of the benchmark results is shown in Table~\ref{tbl:inaturalist}. 

Notice that \fedavgm with \fedvc and a large report goal of $K = 100$ has a 57.2\% accuracy, almost reaching the same level as in centralized learning (57.9\%).  With that said, there is still plenty of room to improve performance with small reporting clients and round budgets. Being able to learn fast with a limited pool of clients is one of the critical research areas for practical visual FL.

\input{floats/fig-landmarks}
\input{floats/tbl-landmarks}

\subsubsection{\LandmarksFed}
The \LandmarksFed dataset comprises 164,172 images for 2,028 landmarks, divided among 1,262 clients. We follow the setup in Section~\ref{sec:importance_reweighting} where we experiment with either training the whole model or fine-tuning the last two layers. Report goal $K=$ \{10, 50, 100\} are used.

Similarly, \fedavgm with the $K=100$ is able to achieve 65.9\% accuracy at 5k communication rounds, which is just 1.2\% off from centralized learning. Interestingly, when we train only the last two layers with FL, the accuracy is as well not far off from centralized learning (39.8\% compared to 40.3\%)

\subsection{Implementation Details} 
\label{sec:implementation}

We use MobileNetV2~\cite{sandler2018mobilenetv2} pre-trained on ImageNet~\cite{deng2009imagenet} for both iNaturalist and \LandmarksFed experiments; for the latter, a 64-dimensional bottleneck layer between the 1280-dimensional features and the softmax classifier. We replaced BatchNorm with GroupNorm~\cite{wu2018group} due to its superior stability for FL tasks~\cite{hsieh2019non}. During training, the image is randomly cropped then resized to a target input size of 299$\times$299 (iNaturalist) or 224$\times$224 (Landmarks) with scale and aspect ratio augmentation similar to~\cite{szegedy2015going}. A weight decay of $4\times10^{-5}$ is applied. 

For CIFAR-10 and CIFAR-100 experiments, we use a CNN similar to LeNet-5~\cite{lecun1998gradient} which has two 5$\times$5, 64-channel convolution layers, each precedes a 2$\times$2 max-pooling layer, followed by two fully-connected layers with 384 and 192 channels respectively and finally a softmax linear classifier. This model is not the state-of-the-art on the CIFAR datasets, but is sufficient to show the relative performance for our investigation. Weight decay is set to $4\times10^{-4}$. 

Unless otherwise stated, the client learning rate is 0.01 and momentum $\beta=0.9$ is used for \fedavgm. The learning rate is kept constant without decay for simplicity. The client batch size is 32 for \LandmarksFed and 64 for others.

%% file: floats/tbl-inaturalist.tex
\begin{table}[t]
\centering
\caption{ \textbf{iNaturalist-\inatuser accuracy.} Numbers reported at fixed communication rounds. $K$ denotes the report goal per round. }
\label{tbl:inaturalist}
\scriptsize
\begin{tabular}{ L{1.25cm}C{1.0cm}C{1.0cm}C{1.0cm}R{1.0cm}R{1.0cm}R{1.0cm} }
\toprule
\multirow{2}{*}[-0.2em]{\textbf{ Method}} & \multirow{2}{*}[-0.2em]{\textbf{ \fedvc}} & \multirow{2}{*}[-0.2em]{\textbf{ \fedimp}} & \multirow{2}{*}[-0.2em]{\textbf{ $K$}} & \multicolumn{3}{c}{\textbf{ Accuracy@Rounds(\%)}} \\ \cmidrule{5-7}
  &   &   &   & \multicolumn{1}{c}{1k} & \multicolumn{1}{c}{2.5k} & \multicolumn{1}{c}{5k} \\ \midrule
\fedavg & \ding{51} & \ding{55} & 10 & 31.3 & 39.7 & 43.9 \\
\fedavg & \ding{51} & \ding{55} & 100 & 36.9 & 46.5 & 51.4 \\ \hline
\fedavg & \ding{51} & \ding{51} & 10 & 30.1 & 41.3 & 47.5 \\
\fedavg & \ding{51} & \ding{51} & 100 & 35.5 & 44.8 & 49.8 \\ \hline
\fedavgm & \ding{51} & \ding{55} & 10 & 37.9 & 43.7 & 49.1 \\
\fedavgm & \ding{51} & \ding{55} & 100 & \textbf{53.0} & \textbf{56.1} & \textbf{57.2} \\ \hline
\fedavgm & \ding{51} & \ding{51} & 10 & 38.4 & 42.1 & 47.0 \\
\fedavgm & \ding{51} & \ding{51} & 100 & 51.3 & 54.3 & 56.2 \\ \midrule
\multicolumn{4}{l}{ Centralized} & \multicolumn{2}{r}{ 57.9} &   \\ \bottomrule
\end{tabular}
\end{table}

%% file: floats/fig-landmarks.tex
\begin{figure}[t]
    \centering
    \includegraphics[width=0.8\linewidth]{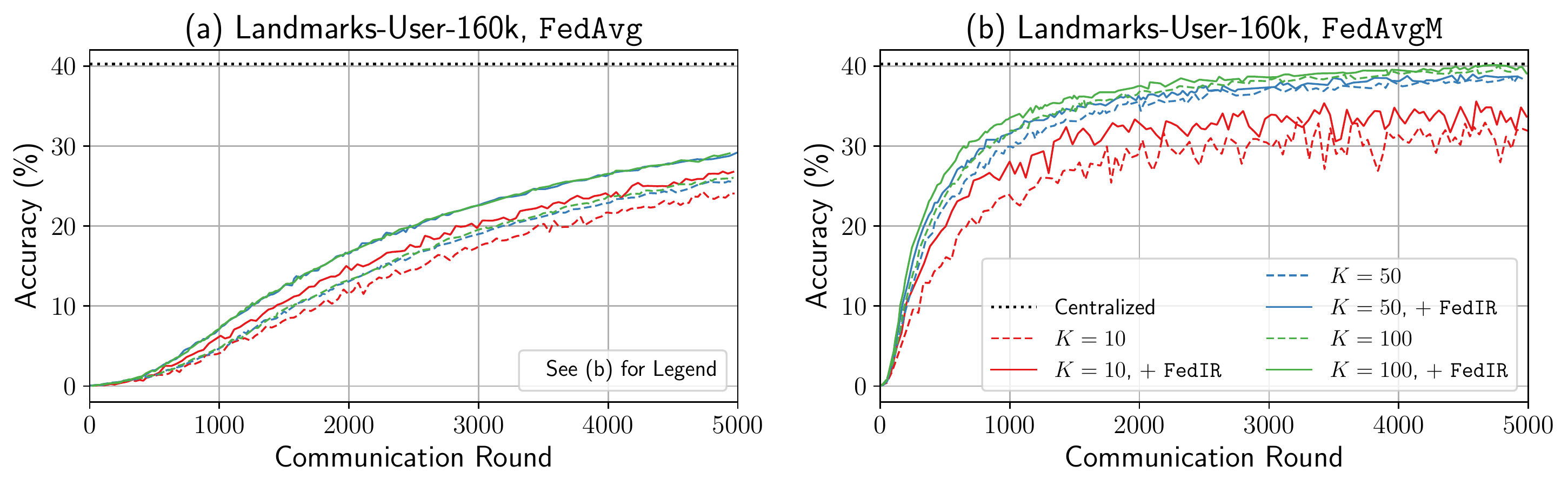}
    \caption{\textbf{\LandmarksFed Learning Curves.} Only the last two layers of the network are fine-tuned. \fedimp is also shown due to its ability to address skewed training distribution as presented in this dataset.}
    \label{fig:landmarks}
\end{figure}

%% file: floats/tbl-landmarks.tex
\begin{table}[t]
\centering
\caption{ \textbf{\LandmarksFed Accuracy.} }
\label{tbl:landmarks}
\scriptsize
\begin{tabular}{ L{1.25cm}C{1.0cm}C{1.0cm}R{1.0cm}R{1.0cm}R{1.0cm}R{1.0cm}R{1.0cm}R{1.0cm} }
\toprule
\multirow{3}{*}[-0.2em]{\textbf{ Method}} & \multirow{3}{*}[-0.2em]{\textbf{ \fedimp}} & \multirow{3}{*}[-0.2em]{\textbf{ $K$}} & \multicolumn{6}{c}{\textbf{ Accuracy@Rounds(\%)}} \\ \cmidrule{4-9}
  &   &   & \multicolumn{3}{c}{\textbf{ Two layers}} & \multicolumn{3}{c}{\textbf{ All layers}} \\ \cmidrule{4-6} \cmidrule(l){7-9}
  &   &   & \multicolumn{1}{c}{1k} & \multicolumn{1}{c}{2.5k} & \multicolumn{1}{c}{5k} & \multicolumn{1}{c}{1k} & \multicolumn{1}{c}{2.5k} & \multicolumn{1}{c}{5k} \\ \midrule
\fedavg & \ding{55} & 10 & 4.2 & 14.6 & 24.6 & 18.2 & 38.1 & 49.7 \\
\fedavg & \ding{55} & 50 & 4.5 & 16.5 & 26.0 & 20.9 & 42.0 & 53.3 \\
\fedavg & \ding{55} & 100 & 4.9 & 16.5 & 26.3 & 21.9 & 42.3 & 53.4 \\ \hline
\fedavg & \ding{51} & 10 & 6.3 & 17.4 & 26.6 & 19.6 & 38.5 & 51.7 \\
\fedavg & \ding{51} & 50 & 7.4 & 19.7 & 28.8 & 26.0 & 45.2 & 55.0 \\
\fedavg & \ding{51} & 100 & 7.2 & 20.1 & 29.0 & 26.5 & 45.7 & 55.2 \\ \hline
\fedavgm & \ding{55} & 10 & 23.0 & 30.1 & 30.8 & 29.4 & 44.1 & 53.7 \\
\fedavgm & \ding{55} & 50 & 29.9 & 36.4 & 38.6 & 55.2 & 62.0 & 64.8 \\
\fedavgm & \ding{55} & 100 & 31.9 & 37.4 & 39.6 & 56.3 & 63.4 & 65.0 \\ \hline
\fedavgm & \ding{51} & 10 & 26.5 & 32.1 & 31.3 & 27.9 & 45.1 & 53.5 \\
\fedavgm & \ding{51} & 50 & 31.6 & 37.5 & 38.9 & 53.1 & 61.6 & 63.2 \\
\fedavgm & \ding{51} & 100 & \textbf{33.7} & \textbf{38.3} & \textbf{39.8} & \textbf{57.7} & \textbf{64.1} & \textbf{65.9} \\ \midrule 
\multicolumn{3}{l}{ Centralized} & \multicolumn{2}{r}{ 40.27} &   & \multicolumn{2}{r}{ 67.05} &   \\ \bottomrule
\end{tabular}
\end{table}

%% file: sections/7_conclusions.tex
\section{Conclusions}

We have shown that large-scale visual classifiers can be trained using a privacy-preserving, federated approach, and highlighted the challenges that per-user data distributions pose for learning. 
We provide two new datasets and benchmarks, providing a platform for other explorations in this space. 
We expect others to improve on our results, particularly when the number of participating clients and round budget is small. There remain many challenges for Federated Learning that are beyond the scope of this paper: real world data may include domain shift, label noise, poor data quality and duplication. Model size, bandwidth and unreliable client connections also pose challenges in practice. We hope our work inspires further exploration in this area.

\section*{Acknowledgements}

We thank Andre Araujo, Grace Chu, Tobias Weyand, Bingyi Cao, Huizhong Chen, Tomer Meron, and Hartwig Adam for their valuable feedback and support.

%% file: sections/_appendix.tex
\appendix

\section{Additional Experiments}
\subsection{Hyperparameter Sensitivity}
\label{sec:hyper}

To study how sensitive the hyperparameter tuning process is to different degrees of non-identicalness in FL settings, we perform experiments on CIFAR-10/100 datasets with a grid of hyperparameters.\footnote{CIFAR experiments in the main text are tuned over the the same grid.}
Following \cite{shallue2018measuring}, we define the effective learning rate for \fedavgm as $\eta_\eff = \eta / \para{1 - \beta}$.
For all values of Dirichlet concentration $\alpha$, we sweep over learning rate $\eta_\eff \in \paral{10^{-3}, 10^{-2.5}, \ldots, 10^0}$ and momentum $1 - \beta \in \paral{10^{-2.5}, 10^{-2}, \ldots, 10^0}$.

\input{floats/fig-hyper}

In Figure~\ref{fig:hyper} we show the effect of using different $\eta_\eff$ on the relative accuracy with each grid point showing the best result over all $\para{\beta, \eta}$ combinations that give the same $\eta_\eff$.

We train for 10k/20k communication rounds with CIFAR-10/100 respectively.

Within each individual contour plot, it can be seen that the accuracy consistently drops with increased non-identicalness, and
the set of hyperparameters yielding high performance becomes smaller.
In general, we find an effective learning rate $\eta_\eff=10^{-2}$ works well in many situations.

Across different report goals $K$, a larger $K$ enables good performance over a wider range of $\eta_\eff$. 
This result is unsurprising, since with more clients reporting in, the server observes more data and hence obtains gradients with less variance.  The number of local epochs does not affect the choice of hyperparameters much in our experiments (see last two rows of Figure~\ref{fig:hyper}).
Interestingly, while CIFAR-10 and CIFAR-100 have different numbers of classes and centralized learning accuracy, they exhibit very similar characteristics in terms of \emph{relative accuracy} (the overall shape of plots in Figure~\ref{fig:hyper} is similar).

\subsection{The Effect of Pretraining}

Pretraining large visual models (e.g., using ImageNet) is very common in centralized training. It is likely to be  even more beneficial in federated settings, where extra computation rounds  could be prohibitively time consuming. In some cases, however, it may be necessary or desirable to train from scratch. In this section, we investigate the feasibility of training large federated visual classification models without pretraining\footnote{Note that in the main text, the smaller CIFAR10/100 experiments are trained from scratch, but the larger iNaturalist and Landmarks experiments use an ImageNet pretrained MobileNetV2.}.

We perform experiments using iNaturalist-\inatgeo{3k} with a combination of settings including the FL algorithm (\fedavg/\fedavgm) and report goal $K$.
Since training from random initialization and from pretrained weights converge to different final test accuracy, we use \textit{relative accuracy} for evaluating FL algorithms' progress relative to the corresponding centralized learning upperbounds.

\input{floats/fig-pretrain}
\input{floats/tbl-pretrain}

From Figure~\ref{fig:pretrain}, we see that FL with pretraining requires orders of magnitude fewer communication rounds for convergence 
and yields higher final relative accuracy than training from scratch.
Table~\ref{tbl:pretrain} further shows the rounds needed to reach 10\%, 50\%, and 90\% relative accuracy.
We see that \fedavgm is able to accelerate convergence significantly, with a report goal $K=100$ it takes 94\% (977 $\rightarrow$ 60) fewer rounds than $\fedavg$ to reach 10\% relative accuracy when starting from pretrained model weights.
We also see that \fedavgm has a much steeper learning curve, reaching 90\% relative accuracy in 6.9$\times$ the rounds needed to reach 10\% (compared to 20$\times$ for $\fedavg$).

Whilst our results suggest that it is possible to train large federated visual classification models from scratch, doing so efficiently and effectively remains an open challenge with room for improvement.

\section{CIFAR-10/100 Dataset Details}
\label{sec:app-cifar}
\subsection{Synthetic Clients with Dirichlet Prior}

To generate non-identical client datasets from CIFAR-10 and CIFAR-100~\cite{krizhevsky2009learning} datasets, we partition each into 100 clients, with 500 training examples each. 
We assume every client $k$ has their data independently drawn from the original dataset according to a multinomial distribution $q_k\para{\cdot}$ of $C$ classes ($q_k\para{y} \geq 0$ and $\sum_y q_k\para{y} = 1$).

To synthesize a population of non-identical clients, we draw a multinomial $\vq_k \sim \dir\para{\alpha \vp}$ from a Dirichlet distribution, where $\vp$ describes a prior class distribution over $C$ classes, and $\alpha > 0 $ is a parameter controlling the \emph{concentration}, or identicalness among all clients. 
$\alpha$ can be used to control the overall homogeneity: $\alpha \rightarrow \infty$ generates clients that are all identical to the prior $\vp$, while $\alpha \rightarrow 0$ generates clients that tend to hold very sparse labels.
After drawing the class distributions $\vq_k$, for every client $k$, we sample training examples from CIFAR-10/100 for each class  according to $\vq_k$ \textit{without replacement}. This is to ensure there are no overlapping examples between any two clients. 

Note that by drawing examples without replacement, towards the end of the assignment process, some subset $\cS$ of classes can be exhausted earlier than other classes, ending up with a shorter list of available classes from which the client synthesis procedure can continue drawing samples. When this happens, we eliminate $\cS$ and enforce the remaining clients to only sample from classes $\paral{1, 2, \ldots, C}\setminus\cS$ with a multinomial distribution 

\begin{equation}
    \tilde q_k \para{y} = 
    \begin{cases}
        0,
            & y \in \cS \\
        q_k \para{y} / \para{1 - \sum_{s \in \cS} q_k\para{s}},
            & y \notin \cS. \\
    \end{cases}
\end{equation}

For CIFAR-10, we use $\alpha \in$ \{100, 10, 1, 0.5, 0.2, 0.1, 0.05, 0\}; for CIFAR-100 we use $\alpha \in$ \{1000, 100, 10, 5, 2, 1, 0.5, 0\}. Summary statistics showing the class count over the client population in both datasets is given in Figure~\ref{fig:cifar-stat}.

\input{floats/fig-cifar-stat}

\section{Experiment Run Time}

The federated learning experiments are carried out by simulation with a cluster of NVIDIA Tesla P100 GPUs in parallel. The experiment run time, while highly variable depending on the experimental setup (model complexity, dataset, local steps $E$, and reporting clients per round $K$), is roughly 0.5 to 2.0 seconds per communication round per reporting client. This amounts to about 9 GPU-days for a run of \LandmarksFed experiment for 5000 rounds with $K=100$.

%% file: floats/fig-hyper.tex
\begin{figure}[h]
    \centering
    \includegraphics[width=\linewidth]{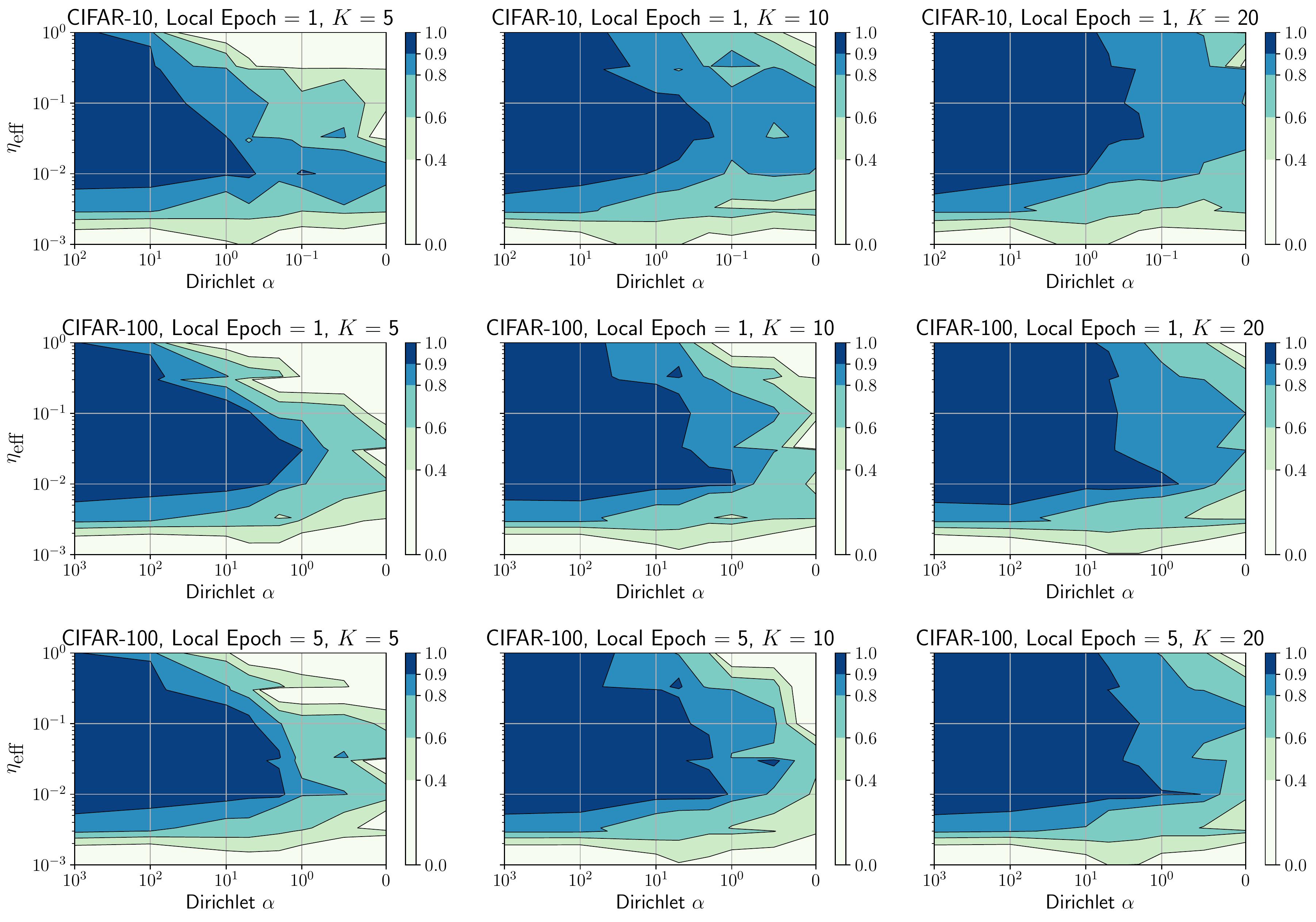}
    \caption{
        \textbf{Relative Accuracy of \fedavgm on CIFAR Datasets.} 
        Darker shades denote regions of higher relative accuracy. 
        $\eta_\eff = \eta / \para{1 - \beta}$ is the effective learning rate, and $K$ is the reporting goal out of 100 clients. 
        Note that data split is increasingly non-identical to the right.
    }
    \label{fig:hyper}
\end{figure}

%% file: floats/fig-pretrain.tex
\begin{figure}[t]
    \centering
    \includegraphics[width=1.0\linewidth]{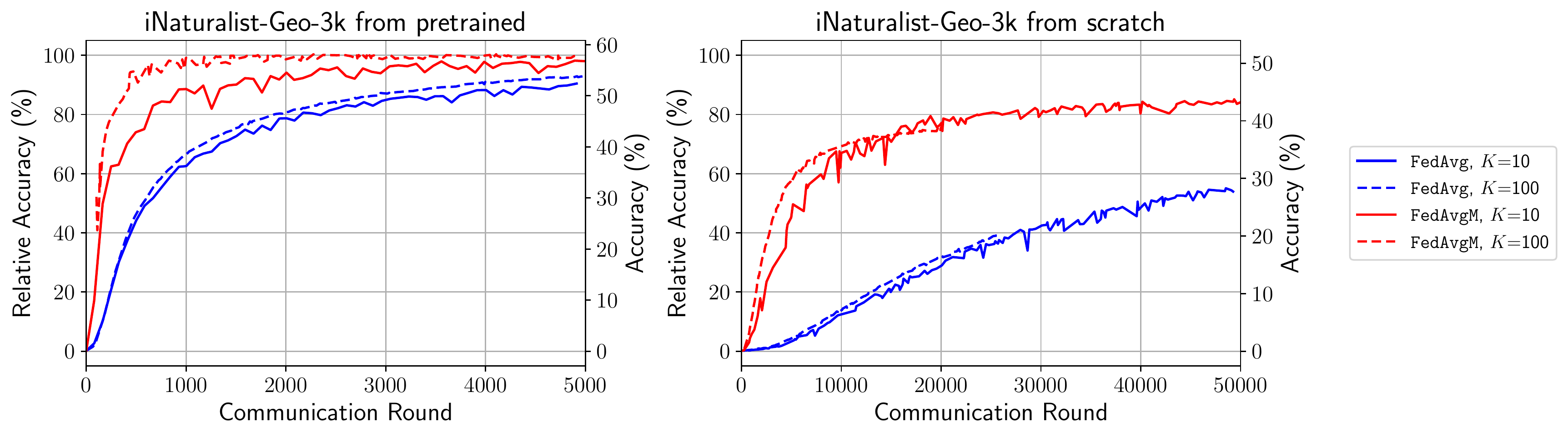}
    \caption{\textbf{Learning Curves from ImageNet Pretraining and from Scratch.} On the left vertical axis is the relative accuracy while on the right is the absolute accuracy. Two plots are rescaled to have the full span of 100\% relative accuracy.}
    \label{fig:pretrain}
\end{figure}

%% file: floats/tbl-pretrain.tex
\begin{table}[t]
\centering
\caption{ \textbf{Communication Rounds to Reach Relative Accuracy.}  Note that models have different centralized learning accuracy (51.4\% from scratch and 57.9\% from pretrained). The multipliers are calculated row-wise, using Rounds@10\% as the baseline. Experiments that do not reach the target relative accuracy even after $t$ rounds is marked $>t$. }
\label{tbl:pretrain}
\scriptsize
\begin{tabular}{ C{1.5cm}L{1.25cm}C{2.0cm}C{1.0cm}R{1.75cm}R{1.75cm}R{1.75cm} }
\toprule
\multirow{2}{*}[-0.2em]{\textbf{ Data}} & \multirow{2}{*}[-0.2em]{\textbf{ Method}} & \multirow{2}{*}[-0.2em]{\textbf{ Initialization}} & \multirow{2}{*}[-0.2em]{\textbf{ $K$}} & \multicolumn{3}{c}{\textbf{ Rounds@Relative Accuracy}} \\ \cmidrule{5-7}
  &   &   &   & 10 \% & 50 \% & 90 \% \\ \midrule
\multirow{8}{*}[-0.3em]{\textbf{ \inatgeo{3k}}} & \fedavg & pretrained & 10 & 165 (1.0$\times$) & 669 (4.1$\times$) & 4912 (29.8$\times$) \\
  & \fedavg & pretrained & 100 & 165 (1.0$\times$) & 567 (3.4$\times$) & 3780 (22.9$\times$) \\ 
  & \fedavgm & pretrained & 10 & 79 (1.0$\times$) & 249 (3.2$\times$) & 1505 (19.1$\times$) \\
  & \fedavgm & pretrained & 100 & \textbf{60} (1.0$\times$) & \textbf{116} (1.9$\times$) & \textbf{420} (6.9$\times$) \\ \cline{2-7}
  & \fedavg & scratch & 10 & 9005 (1.0$\times$) & 39236 (4.4$\times$) & $>$ 50k \\
  & \fedavg & scratch & 100 & 7793 (1.0$\times$) & $>$ 20k & $>$ 20k \\
  & \fedavgm & scratch & 10 & 1463 (1.0$\times$) & 5788 (4.0$\times$) & $>$ 50k \\
  & \fedavgm & scratch & 100 & 977 (1.0$\times$) & 3733 (3.8$\times$) & $>$ 20k \\ \bottomrule
\end{tabular}
\end{table}

%% file: floats/fig-cifar-stat.tex
\begin{figure}[t]
    \centering
    \begin{subfigure}[t]{0.5\linewidth}
        \centering
        \includegraphics[width=\linewidth]{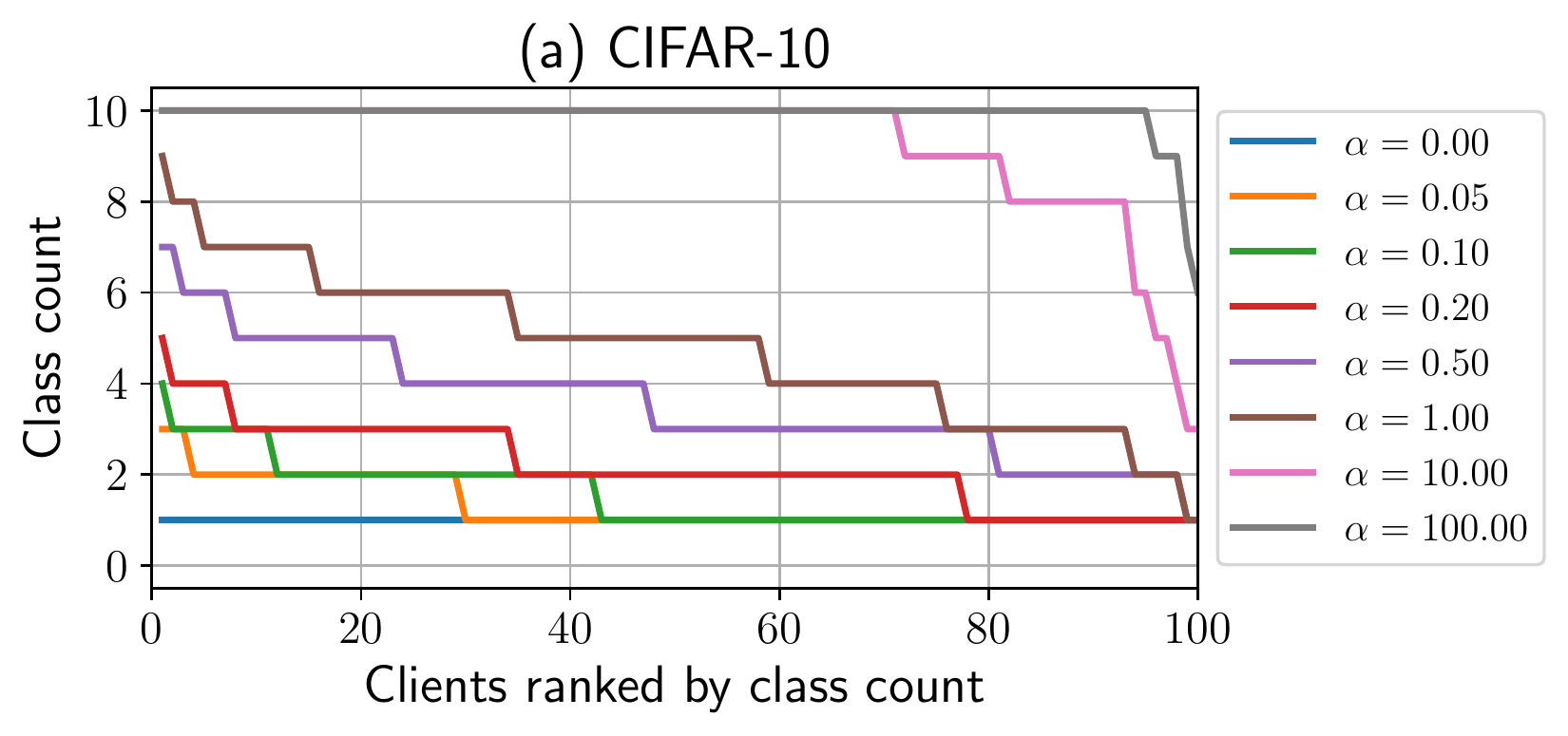}
    \end{subfigure}%
    ~
    \begin{subfigure}[t]{0.5\linewidth}
        \centering
        \includegraphics[width=\linewidth]{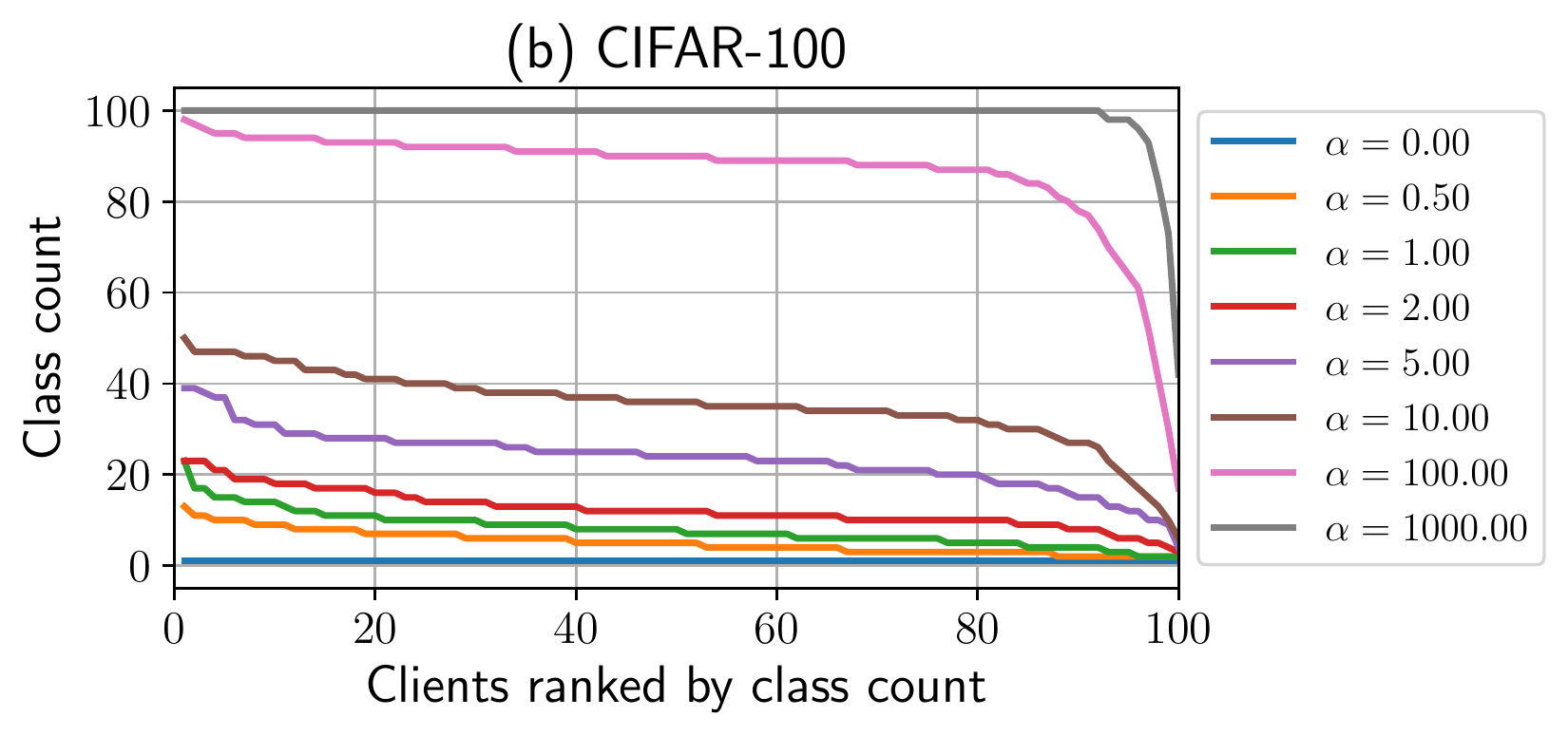}
    \end{subfigure}
    \caption{\textbf{CIFAR-10/100 Distribution.} Each curve represents the class counts of clients within a data partitioning synthesized using a Dirichlet concentration parameter $\alpha$.
    }
    \label{fig:cifar-stat}
\end{figure}